\documentclass[10pt,tikz]{article}

\usepackage[toc,page]{appendix}
\usepackage[official]{eurosym}
\usepackage{amsmath}
\usepackage{amssymb}
\usepackage{array}
\usepackage{chngcntr}
\usepackage{courier}
\usepackage{enumitem}
\usepackage{float}
\usepackage[edges]{forest}
\usepackage{graphics}
\usepackage{graphicx}
\usepackage[linktocpage]{hyperref}
\usepackage{mathtools}
\usepackage{mdframed}
\usepackage{parskip}
\usepackage{pdflscape}
\usepackage{pifont}
\usepackage[section]{placeins}
\usepackage{scrextend}
\usepackage{sectsty}
\usepackage{siunitx}
\usepackage{tabularx,colortbl}
\usepackage{tcolorbox}
\usepackage{titling}
\usepackage{tocloft}
\usepackage{url}
\usepackage{xcolor}
\hypersetup{
	colorlinks,
	linkcolor = {red!50!black},
	citecolor = {blue!50!black},
	urlcolor  = {blue!80!black}
}

\usepackage[utf8]{inputenc}
\usepackage{menukeys}

\DeclareMathOperator{\atantwo}{atan2}
\DeclarePairedDelimiter\abs{\lvert}{\rvert} 

\newcolumntype{M}[1]{>{\centering\arraybackslash}m{#1}}
\newcolumntype{N}{@{}m{0pt}@{}}

\newcolumntype{C}{c@{\hspace{1mm}}}
\newcolumntype{L}{l@{\hspace{1mm}}}
\newcolumntype{R}{r@{\hspace{1mm}}}

\graphicspath{{figs/part_1/}}

\usepackage[lmargin=2cm,rmargin=2cm,tmargin=2cm,bmargin=2cm]{geometry}
\usepackage[font=footnotesize, labelfont=bf]{caption}

\counterwithin*{section}{part}

\partfont{\LARGE}

\title{\textbf{Robust Video-Based Eye Tracking Using Recursive Estimation of Pupil Characteristics} \vspace{0.5cm}} 
\author{%
	\textsc{Terence Brouns} \\[1ex] 
	\normalsize Radboud University, the Netherlands \\ 
	\normalsize \href{mailto:t.s.n.brouns@gmail.com}{t.s.n.brouns@gmail.com} 
}
\date{\vspace{-5ex}}

\begin{document}
	
	\maketitle
	\section*{Abstract}
	
	Video-based eye tracking is a valuable technique in various research fields. Numerous open-source eye tracking algorithms have been developed in recent years, primarily designed for general application with many different camera types. These algorithms do not, however, capitalize on the high frame rate of eye tracking cameras often employed in psychophysical studies. We present a pupil detection method that utilizes this high-speed property to obtain reliable predictions through recursive estimation about certain pupil characteristics in successive camera frames. These predictions are subsequently used to carry out novel image segmentation and classification routines to improve pupil detection performance. Based on results from hand-labelled eye images, our approach was found to have a greater detection rate, accuracy and speed compared to other recently published open-source pupil detection algorithms. The program's source code, together with a graphical user interface, can be downloaded at \url{https://github.com/tbrouns/eyestalker}. 
	
	\textbf{\textit{keywords:}} eye tracking, video-oculography, pupil detection, open source, algorithm, psychophysics, methodology
	
	\section{Introduction}
	
	The act of measuring the movement of the eye, known as eye tracking, enjoys a broad range of applications across various disciplines, from neuroscience and psychology to market research and industrial engineering \cite{duchowski2002}. Within neuroscience, eye tracking can serve as a method to diagnose neurological problems such as dyslexia, schizophrenia, Alzheimer's disease, attention deficit hyperactivity disorder and autism \cite{dolezal2015}, but can also be used as a research tool to study the visual, auditory \cite{volck2015} and vestibular systems \cite{allison1996}. Distinct eye tracking techniques have been developed over the years, each with their own set of advantages and disadvantages. In electro-oculography (EOG) small differences in the electric potential on the skin, caused by the retinal polarity, are measured with surface electrodes placed around the eye, which give an estimate of the eye position. The method is often employed in sleep studies, because it works when the eye is closed and one can measure for long stretches at a time without user discomfort. However, it suffers from poor accuracy and unreliability of vertical eye movement recordings \cite{heide1999}. Far greater resolution is achieved with the search coil technique, which is considered the gold standard for eye tracking \cite{geest2002}. It uses a small copper coil embedded in a contact lens that is placed on the eye. A voltage is induced in the coil in the presence of an external oscillating magnetic field. The voltage amplitude gives a measure of the eye position, because it is proportional to the orientation of the coil in the magnetic field. Like in EOG, this operational mechanism allows eye movement recording even when the eye is closed. An obvious drawback of the search coil method is the invasive nature of mounting and removing the lens. Furthermore, corneal irritation and erosion can occur in long recording sessions from wearing the coil \cite{heide1999}.
	
	A compromise between the high accuracy of the search coil technique and the non-invasiveness of EOG can be offered by video-based eye tracking, also known as video-oculography (VOG). In VOG, the eye is tracked using a camera combined with computer vision algorithms. We can distinguish between two different approaches, which differ in how close the camera is removed from the eye and therefore require different image processing techniques. Much published work has focussed on developing eye tracking algorithms for camera systems that capture the whole head (e.g. webcam-based). These systems are particularly useful in human-computer interaction and are noted for their ease of use, but severely lack in accuracy, which makes them unsuitable for research purposes where eye tracking accuracy is crucial, such as saccade analysis. In order to compete with the search coil technique in terms of spatial resolution, the camera must be brought much closer to the eye. Generally speaking, eye movement is tracked in these close-up images of the eye by finding the pupil position through feature detection. Detection is aided by illumination of the eye with an infra-red (IR) light emitting diode (LED), which turns the pupil much brighter or darker, depending on whether the IR LED is close to or away from the camera's optical axis \cite{morimoto2000}, making the pupil a more clearly discernible feature in the image. The use of the IR light also means that the system is functional in the dark, which is often a requirement in experimental research. 
	
	Commercially available state-of-the-art eye trackers with the aforementioned VOG set-up are generally expensive, upwards of \$10.000. In addition to their price tag, another downside of commercial eye trackers is their use of proprietary software, which often prevents any customization to fit the specific needs of the user. Fortunately, a number of developers have been spurred on to produce low-cost hardware alternatives \cite{abbott2012}\cite{adiba2016}\cite{kim2014}\cite{li2006}\cite{mantiuk2012}\cite{putra2013}\cite{schneider2011}, publish their own sophisticated pupil detection algorithms \cite{li2005}\cite{lin2010}\cite{long2007}\cite{swirski2012}\cite{zhu1999} and make open-source VOG software publicly available \cite{Fuhl2015}\cite{Kassner2014}\cite{zimmermann2016}. Unfortunately, few implementations are suited for use in a wide variety of psychophysical experiments, which require both precision and speed. As technology advances, high-speed cameras with high-resolution sensors will become more affordable. This cutting-edge hardware should be accompanied by the appropriate software, capable of fully capitalizing on the improved camera features. 
	
	Here, we introduce an open-source pupil detection algorithm that aims to exploit the high frequency nature of many modern cameras ($\geq 200$ Hz). The method is based on the notion that the characteristics of the pupil can only change a relatively small amount between consecutive frames when recording at a high frame rate. This allows us to make accurate predictions about the pupil's characteristics for the next frame, where our search will be limited to features in the image that match these predictions. This method contrasts with many published pupil tracking algorithms where each camera frame is effectively treated as being independent from the one that preceded it. An advantage of algorithms that are built on this premise is their universal applicability. They work equally well with different cameras operating at varying speeds and on random collections of eye images. However, this property becomes irrelevant when users only utilize such algorithms in combination with high-speed cameras. The pupil detection method that we present in this text has been specifically tailored for individuals who use these types of powerful cameras under more ideal conditions (e.g. a controlled laboratory setting), giving them the opportunity to take greater advantage of their hardware and environment to improve eye tracking performance.
	
	\section{Methods}
	
\begin{figure}[ht]
		\vspace*{-1.0cm}
		\centering
		\includegraphics[width=\textwidth]{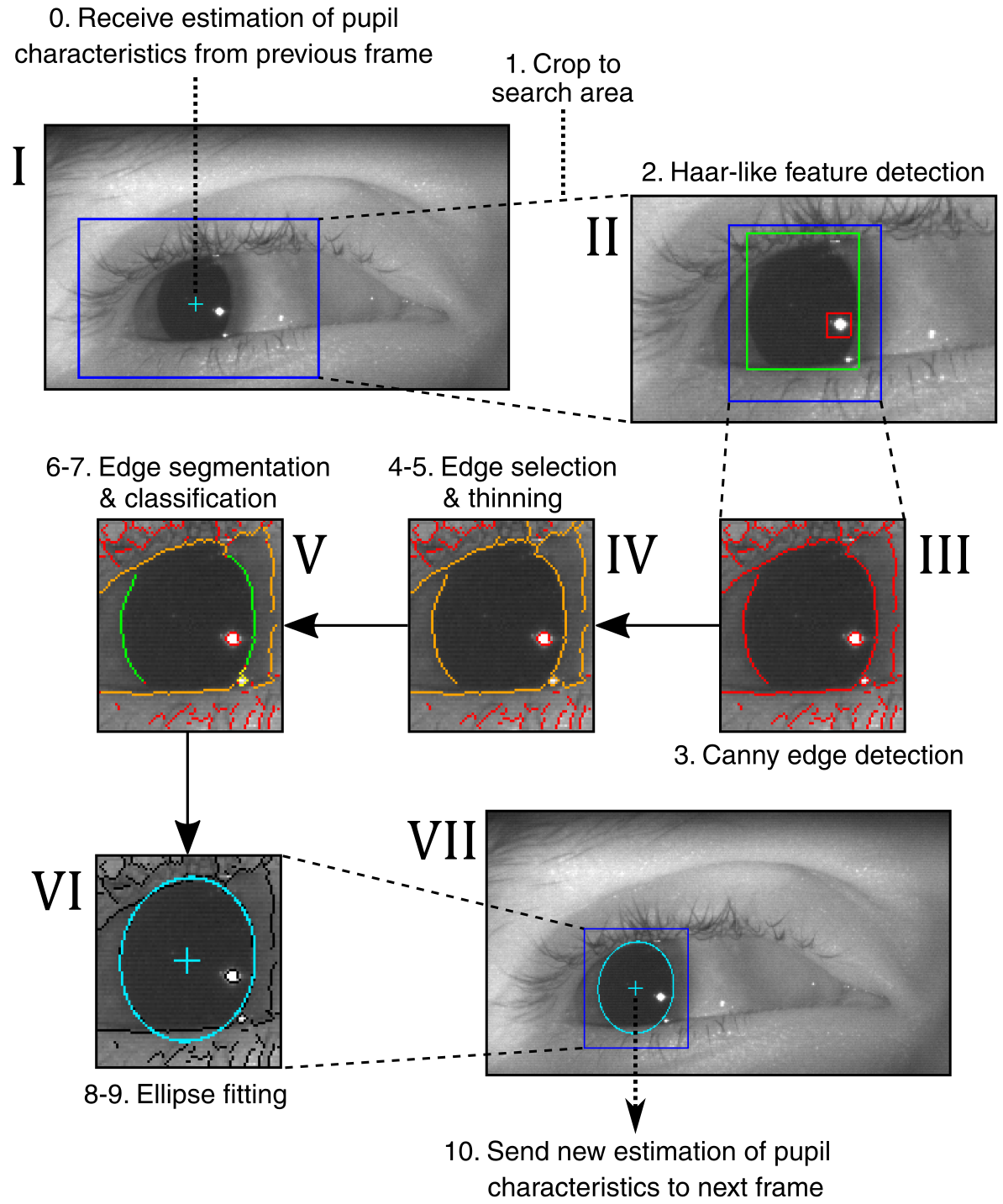}
		\caption{Example of an image being sent through the pupil detection pipeline. The 0 to 10 labels refer to the processing steps as given in the main text. The Roman numerals indicate various explanatory images: \textit{I.} Raw image input in which the pupil position prediction is marked (teal cross). Position prediction is used to obtain the rectangular search area (blue outline). \textit{II.} Resulting image after cropping to search area. Detected position of corneal reflection (red rectangle) is accounted for during Haar-like feature detection of pupil (green rectangle). \textit{III.} Canny edge detection is performed in an area (blue rectangle in \textit{II}) whose dimensions and position are determined by combination of Haar-like feature detection and predicted pupil characteristics. Detected Canny edges are shown in red. \textit{IV.} Sub-set of Canny edges is selected (orange) based on their position. Remaining edges are discarded. \textit{V.} Selected edges are segmented if required. We then choose edge sections that belong to the pupil-iris contour (green), after classification. \textit{VI.} One or more ellipses are fitted on pupil-iris contour edges. Most optimal ellipse fit is accepted (teal), depending on predictions of pupil characteristics. Centre of chosen ellipse (cross) serves as pupil position. \textit{VII.} Ellipse fit characteristics are used to update pupil predictions for next frame.}
		\label{fig:detection_method}
	\end{figure}

	The pupil detection algorithm, which is illustrated in Figure \ref{fig:detection_method}, works by performing a number of distinct processing tasks in the following order:
	
	\begin{enumerate} \itemsep0em
		\setcounter{enumi}{-1}
		\item	Receive predicted pupil characteristics for current frame
		\item	Crop image to smaller search area depending on predicted size, shape and position of pupil
		\item	Update predicted pupil position following object recognition of the pupil through Haar-like feature detection, after removing corneal reflection interference
		\item	Detect all object boundaries, i.e. edges, in area around new position estimate by identifying brightness discontinuities with Canny edge detection   
		\item	Select sub-set of detected edges that are at an acceptable distance from the position estimate 
		\item	Thin Canny edges to minimum thickness using morphological operations
		\item	Segment edges at transition point between distinct features in the image
		\begin{enumerate}
			\item	Choose non-branching path in edges that comes closest to predicted pupil circumference
			\item	Split edges at points of extremely high or low curvature
			\item	Reduce length of edges if they exceed pupil circumference estimate
		\end{enumerate}
		\item	Classify edges into two classes: pupil contour edges and non-pupil contour edges
		\item	Fit one or more ellipses on edges categorized as pupil contour edges
		\item	Choose most optimal fit based on pupil characteristic predictions 
		\item	Calculate new predictions for next frame
	\end{enumerate}
	
	\FloatBarrier
	
	The algorithm relies on various parameters to carry out these steps, some of which depend on the image size and frame rate of the camera. The default parameter values given in the text are based on a reference eye image size of around 400 $\times$ 200 pixels and a frame rate of 250 Hz. Using these reference values, a few parameters should be scaled down or up for other image sizes or sampling rates. Furthermore, we will sometimes refer to a data set from which certain information has been extracted. This data set is a collection of more than 400,000 close-up images of the left eye of 12 different individuals, which were taken over the course of around 100 saccadic eye movements made by each person in various directions, while they were seated in a dark room. The pictures were made by a UI154xLE-M camera from IDS Imaging Development Systems (Obersulm, Germany), using infra-red illumination and with the reference camera settings specified earlier. 
	
	\FloatBarrier
	
	\subsection{Feature value prediction}
	
	The strength of the pupil detection algorithm is drawn from the use of predictions that are made about certain characteristics of the pupil in each camera frame. Consequently, we need a method that is able to make these predictions as accurately as possible, without being computationally demanding and the need to set many different parameters. For online estimation, a recursive estimation method is preferred over batch estimation, because it only requires the measurement from the current frame and the prediction from the previous one, which contrasts with batch estimation where significantly more data has to be processed to make the prediction. A popular type of recursive estimator is the Kalman filter, which has been extensively used in computer vision research \cite{wren1997} and has already been applied to eye tracking in the past \cite{abd2002} \cite{chi2014} \cite{zhang2006}, although limited to whole-face eye tracking. To my knowledge, this is the first application of recursive estimation in eye tracking using close-up images of the eye. Based on this previous work, the Kalman filter may seem like an obvious choice for our recursive estimator as well. However, because we are interested in estimating many different variables, the use of the Kalman filter is made difficult due to the exponentially increased complexity of setting the measurement and process noise covariances, which are integral to the functionality of the filter. Optimal parameter values would have to be found through tuning \cite{welch95}, a process that most likely has to be repeated for each new camera set-up. Since we want our algorithm to be easily implemented in other systems, we have opted to develop a more basic recursive estimation method, which may lack the accuracy of the Kalman filter, but requires fewer parameters to be set. Using our recursive estimator, we keep track of the pupil features that are listed in Table \ref{table:feature_variables}. After the pupil has been detected in a particular camera frame, each one of these features is updated with the following general formula:
	\begin{equation} \label{eq:feature_update}
	\hat{f}_{n+1} = \hat{f}_{n} + \alpha\Delta f_{n} + c_{n}p_{n}
	\end{equation}
	Here, $\hat{f}_{n}$ is the predicted feature value for the current frame $n$. This value is updated to obtain the prediction for the next frame, $\hat{f}_{n+1}$, by adding the difference $\Delta f_{n}$ between the measured and the predicted value (the prediction error), plus a momentum term, $p_{n}$. In every frame where we detect the pupil, we will possess a measured feature value $f_{n}$, which is used to calculate $\Delta f_{n}$ through: 
	\begin{equation}
	\Delta f_{n} = f_{n} - \hat{f}_{n}
	\end{equation}
	
	\begin{table}
		\centering
		\begin{tabular}{r c l}
			
			\textbf{Estimated features} 
			& \textbf{Symbol} 
			& \textbf{Description} \\ 
			
			Position
			& $\hat{s}$ 
			& Position of the centre of the pupil in Cartesian coordinates \\
			
			Circumference 
			& $\hat{C}$ 
			& Circumference of pupil-iris boundary \\
			
			Aspect ratio
			& $\hat{AR}$
			& Ratio between pupil major and minor axes \\
			
			Width
			& $\hat{W}$  
			& Width of pupil bounding box \\
			
			Height 
			& $\hat{H}$ 
			& Height of pupil bounding box \\
			
			Angle
			& $\hat{\theta}$ 
			& Pupil rotation angle \\
			
			Brightness/intensity
			& $\hat{I}$ 
			& Grey-scale value of the inner pupil-iris boundary \\ 
			
			Radial gradient
			& $\hat{G_{r}}$ 
			& Radial image gradient of the pupil-iris boundary \\
			
			Curvature
			& $\hat{\kappa}$ 
			& Signed curvature of pupil-iris boundary \\
			
			
		\end{tabular}
		\caption{Pupil characteristics that are estimated in successive frames with the recursive estimator.}
		\label{table:feature_variables}
	\end{table}
	
	The value of $\Delta f_{n}$ is modified by the gain factor $\alpha$ in equation \ref{eq:feature_update}, which is a constant value between 0 and 1. The larger the gain, the more the estimation is based on the current measurement $f_{n}$. The value that we choose for $\alpha$ should depend on the feature in question. A feature such as pupil position should be updated relatively quickly, because it can shift rapidly on short timescales, requiring a larger $\alpha$.  Significant changes in the pupil’s size and shape, on the other hand, generally occur on longer timescales, allowing for a smaller gain, which helps reduce the influence of noise. If we were to just use $\Delta f_{n}$ to update $f_{n}$ then this approach would fail when the feature value monotonically increases, because our prediction will lag behind. For this reason, a momentum term $p_{n}$ has been added in equation \ref{eq:feature_update}, where $p_{n}$ is updated in every frame by:
	\begin{equation} \label{eq:momentum_update}
	p_{n+1} = p_{n}+ \alpha(\Delta f_{n} - p_{n}) 
	\end{equation}
	
	What $p_{n}$ tries to approximate is the prediction error $\Delta f_{n}$ of the feature. Ideally, $p_{n}$ causes $\Delta f_{n}$ to drop to zero, resulting in accurate predictions.  The rate of change of $p_{n}$ is also reduced by the factor $\alpha$, with the same $\alpha$ value being used for both the prediction error and momentum. Furthermore, $p_{n}$  is modified by an additional factor $c_{n}$ in equation \ref{eq:feature_update}. This quantity gives a measure of the certainty of our prediction and plays an important role in other parts of the algorithm as well. This factor is added to the equation to avoid erratic behaviour when little information about the pupil is available. For example, when the pupil momentarily disappears from view because the eye is closed, we are unsure of its position and our prediction will most likely have a large error, $\Delta f_{n} \gg 0$. So when the pupil appears once again, $p_{n}$ will approximate this large error, resulting in an excessive rate of change, causing $\hat{f}_{n}$ to overshoot the actual value for the feature. To avoid this behaviour, we control the influence of momentum with the degree of certainty $c_{n}$ of our prediction, which is a value between 0 (low) and 1 (high). The value for $c_{n}$ is altered according to the rate of change of the feature. If the rate falls within an acceptable range, $c_{n}$ increases, otherwise it decreases. This is given by:
	\begin{equation}
	\Delta c_{n} = 1 - \frac{2} {1 + e^{k(\delta_{n} - \delta'_{\theta})}}
	\end{equation}
	
	Where $\Delta c_{n}$ indicates the change in certainty, which is calculated using a generalized logistic function that limits $\Delta c_{n}$ to values between $+1$ and $-1$.  The variable $\delta_{n}$ tell us how much the feature value has shifted from one frame to the next. This quantity differs from $\Delta f_{n}$, because $\delta_{n}$ can be a relative change and is always positive, whereas $\Delta f_{n}$ is always an absolute change and can be negative or positive. The constant $\delta'_{\theta}$ is an upper limit of typical $\delta_{n}$ values and is determined empirically (see further down). If $\delta_{n}$ is below this threshold value of $\delta'_{\theta}$, the function will return a positive number, causing $c_{n}$ to increase. Alternatively, $c_{n}$ decreases when $\delta_{n} > \delta'_{\theta}$. The steepness $k$ of the curve  is determined by:
	\begin{equation}
	k= \log\frac{\frac{1}{a} - 1}{b\delta'_{\theta}}
	\end{equation}
	
	Where we set $a = 0.99$ and $b = 0.50$, which translates to the logistic function reaching 99\% of its maximum value at $\delta_{n}=\frac{1}{2}\delta'_{\theta}$. These parameters are set in this way, because the certainty measure should not be biased towards ever smaller $\delta_{n}$ values, but should view all $\delta_{n}$ values that are physiologically feasible as equivalent. For example, how certain we are about the pupil’s position during a fixation period should not automatically be greater than our level of certainty about its position during a saccadic eye movement. Both can realistically occur so should be treated as equal in our certainty calculation. We update the certainty $c_{n}$ by first computing an intermediate value: 
	\begin{equation} \label{eq:certainty_update}
	c'_{n+1} = c'_{n} + \alpha \alpha_c \Delta c_{n} 
	\text{  with  }   
	c'_{n} = 
	\begin{cases}
	1  & \text{if } c'_{n} > 1\\
	0  & \text{if } c'_{n} < 0\\
	\end{cases}
	\end{equation}
	
	Where $\Delta c_{n}$ is modified with the product of $\alpha$ and a new gain factor $\alpha_{c}$. The certainty value that we use in equation \ref{eq:feature_update} is calculated through another logistic function:
	\begin{equation}
	c_{n} = \frac{1}{1 + e^{-\tau(c'_{n}-\frac{1}{2})}}
	\end{equation}
	
	The use of the logistic function has two purposes. First, the function bounds $c_{n}$ in the 0 to 1 range. Second, the function imposes some latency for changes in certainty from the limits, which is controlled by the constant $\tau$ factor (set to 10 by default). As a consequence, multiple precise measurements are required before the certainty starts to significantly increase from its minimum value, or multiple imprecise or non-detections before certainty drops down from the maximum. We essentially create three different states the algorithm can attain: a focussed state when certainty is high ($c \approx 1$), an exploratory state when it is low ($c \approx 0$), and a narrow transitional state in between. One reason for this design is that extreme values of $p_{n}$ are effectively ignored when the algorithm is still in the exploratory state. Another reason is that an all-or-nothing scheme is generally preferred for parts of the algorithm that rely on $c_{n}$ to modify specific parameters. This has the effect that some functions are turned on and off depending on the state the algorithm is in, which can boost performance. 
	
	In order to avoid having to specify a different set of parameters for each feature, we limit our features to two classes, namely position (S) and appearance (A). The position class only has the position prediction $\hat{s}$ as its member. All other features belong to the appearance class. This particular classification is chosen due to the different timescales these features change at, as mentioned before. The same $c_{n}$ and $\alpha$ values are shared between members of each class. We use $c_{S}$ and $\alpha_{S}$ for position, while  $c_{A}$ and $\alpha_{A}$ are used for appearance (with $\alpha_{S} = 0.75$ and $\alpha_{A} = 0.40$). For the position class, $\delta_{n}$ is equal to the absolute change in position (i.e. displacement). For the appearance class, only changes in circumference, $C$, and aspect ratio, $AR$, are considered when calculating $\Delta c_{n}$. The average $\Delta c_{n}$ value between both is used to update $c_{A}$. We only consider these two features, because it can already be assumed that a given measurement is accurate if the measured size and shape of the pupil changes little between consecutive frames, making it redundant to check the change of other features (e.g. brightness) as well.  The various $\delta_n$ values are given by:
	\begin{align}
	\delta_{s}  &= \sqrt{(x_{n} - \hat{x}_{n})^2 + (y_{n} - \hat{y}_{n})^2} \\
	\delta_{C}  &= \frac{\abs{ C_{n} - \hat{C}_{n}}}{\max(C_{n},\hat{C}_{n})} \label{eq:change_C} \\
	\delta_{AR} &= \ \abs{ AR_{n} - \hat{AR}_{n} } \label{eq:change_AR}
	\end{align}
	
	To obtain the threshold $\delta'_{\theta}$ values for $\delta_{s}$, $\delta_{AR}$ and $\delta_{C}$, we look at our data set to see what kind of values we typically obtain for these quantities. For each $\delta'_{\theta}$ type, we choose a threshold at which less than 1\% of measurements have a $\delta_{n}$ value that is larger than $\delta'_{\theta}$. We also determine another threshold, $\delta''_{\theta}$, which is similar to $\delta'_{\theta}$, but uses 0.1\% of measurements as the benchmark instead of 1\%. For our camera set-up, we set the lower thresholds to $\delta'_{\theta,s} = 3$, $\delta'_{\theta,C} = 0.03$, $\delta'_{\theta,AR} = 0.03$ and the upper thresholds to $\delta''_{\theta,s} = 6$,  $\delta''_{\theta,C} = 0.12$, $\delta''_{\theta,AR} = 0.09$. When $\delta_{n} > \delta''_{\theta}$, we make the assumption that the measurement is very likely to have been significantly influenced by noise effects and is not a true physiological result. The $\delta''_{\theta}$ parameter is used at a later point. Both $\delta'_{\theta}$ and $\delta''_{\theta}$ need to be altered according to the frame-rate of the camera. The thresholds can be decreased when using a faster camera and should be increased with slower cameras.
	
	Up until this point, we have not dealt with the issue of non-detections, i.e. frames in which the algorithm has not been able to find the pupil. In the case of a non-detection, measures of $f_{n}$ and $\delta_{n}$ are not available. This stops us from updating our feature predictions and certainty using equations \ref{eq:feature_update} and \ref{eq:certainty_update}.  We deal with this problem in our certainty calculation by giving $\Delta c_{n}$ the minimum value of –1 when a non-detection occurs. For the feature predictions, this issue is resolved by introducing the average feature variable $\bar{f}_{n}$. This average feature variable is a quantity that our prediction $\hat{f}_{n}$ can fall back on when no immediate information about the pupil is available. We only calculate $\bar{f}_{n}$ for features belonging to the appearance class, since an average position is not a meaningful quantity in this context. For position, we instead rely on the detection algorithm to supply an estimation of the position prediction $\hat{s}_{n}$ during non-detections (see section \ref{sec:approximate_detection}). We update $\hat{f}_{n}$ during a non-detection according to:
	\begin{equation}
	\hat{f}_{n+1} = \hat{f}_{n} + \alpha_{A} \Delta \hat{f}_{n} + c_{A,n} p_{n}	
	\end{equation}
	\begin{equation}
	\Delta \hat{f}_{n} = \bar{f}_{n} - \hat{f}_{n}
	\end{equation}
	
	\noindent
	The momentum term $p_{n}$ decays to zero by setting $\Delta \hat{f}_{n}$ to zero in equation \ref{eq:momentum_update}, leaving:
	\begin{equation}
	p_{n+1} = (1 - \alpha_{A})p_{n} 
	\end{equation}
	
	\noindent
	We update $\bar{f}_{n}$ in every frame in a similar fashion to $\hat{f}_{n}$, but without using the momentum term:
	\begin{equation}
	\bar{f}_{n+1} = \bar{f}_{n} + \alpha_{mean} \Delta \bar{f}_{n}
	\end{equation}
	\begin{equation}\label{eq:prediction_error_avg}
	\Delta \bar{f}_{n} = \hat{f}_{n} - \bar{f}_{n}
	\end{equation}
	
	The gain factor of $\alpha_{mean}$ should be much smaller than $\alpha_{A}$ in order for $\bar{f}_{n}$ to keep track of an overall mean of $f_{n}$ (we use $\alpha_{mean} = 0.005$). During a non-detection, $\hat{f}_{n}$ in equation \ref{eq:prediction_error_avg} is replaced by a typical value that is initially assigned to $\hat{f}_{n}$ and $\bar{f}_{n}$ at the start of detection. The end result is that in every frame we have a prediction $\hat{f}$ for each type of feature, together with a measure $c$ of how certain we are about that prediction. These quantities play a role in almost every part of the pupil detection algorithm.
	
	\FloatBarrier
	
	\subsection{Search area}
	
	The first application of our predictions and certainties is in reducing the processing area for pupil detection. By recognizing that the pupil can only translate and transform a limited amount between consecutive frames, we can narrow our search of the pupil to an area that is much smaller than the total size of the image. Obviously, the width $W_{AOI}$ and height $H_{AOI}$ of this area of interest (AOI) still need to be at least as large as the predicted width $\hat{W}$ and height $\hat{H}$ of the pupil, so we set:
	\begin{equation}
	W_{AOI} = \Delta L + \hat{W}
	\end{equation}
	\begin{equation}
	H_{AOI} = \Delta L + \hat{H}
	\end{equation}
	
	Here, $\Delta L$ is some additional length that is added to our prediction for the pupil size to ensure that the pupil falls wholly in the AOI. This length is determined by how much the pupil could potentially move and resize between two successive frames:
	\begin{equation}
	\Delta L = \frac{\hat{C}\delta_{\theta,C}}{\pi} + 2 \delta_{\theta,s}
	\end{equation}
	
	\noindent
	The variables $\delta_{\theta,s}$ and $\delta_{\theta,C}$ are derived from the threshold values $\delta''_{\theta}$ in the following way:
	\begin{alignat}{2}
	\delta_{\theta,s} &= (1 - c_{S}) (\delta_{max,s} - \delta''_{\theta,s}) &&+ \delta''_{\theta,s} \\
	\delta_{\theta,C} &= (1 - c_{A}) (\delta_{max,C} - \delta''_{\theta,C}) &&+ \delta''_{\theta,C} \label{eq:delta_circumference}
	\end{alignat}
	
	The degree of certainty modifies the size of the AOI, with greater $c$ values resulting in smaller processing areas, up to a minimum value of $\delta_{\theta}$ and a maximum value of $\delta_{max}$. These maximum values are a type of theoretical upper limit and are given by:
	\begin{align}
	\delta_{max,s} &= \max(W_{img} - \hat{W}, H_{img} - \hat{H}) \\
	\delta_{max,C} &= \max(\frac{C_{max} - \hat{C}}{C_{max}}, \frac{\hat{C} - C_{min}}{\hat{C}}) 
	\end{align}
	
	Where $W_{img}$ and $H_{img}$ respectively are the width and height of the input image. The maximum circumference change is calculated from the smallest possible pupil circumference ($C_{min}$) and the largest ($C_{max}$), which are empirically determined for our set-up.
	
	\FloatBarrier
	
	\subsection{Approximate detection}
	\label{sec:approximate_detection}
	
	For several parts of the algorithm, it is essential that a rough estimate of the pupil position is available. When $c_{S}$ is low, we cannot completely rely on $\hat{s}$ to provide this approximation, because it is most likely inaccurate. In that case, $\hat{s}$ is re-evaluated by convolving the image with a Haar-like feature detector \cite{viola2001}. This type of detector works by moving a Haar-like feature over the image and calculating the difference in the sum of all pixel values between the dark and light rectangular regions. This detection method is made computationally efficient by first calculating the integral image, which ensures $O(n)$ performance with respect to the number of input pixels.
	
\begin{figure}[ht]
		\centering
		\includegraphics[width=0.6\textwidth]{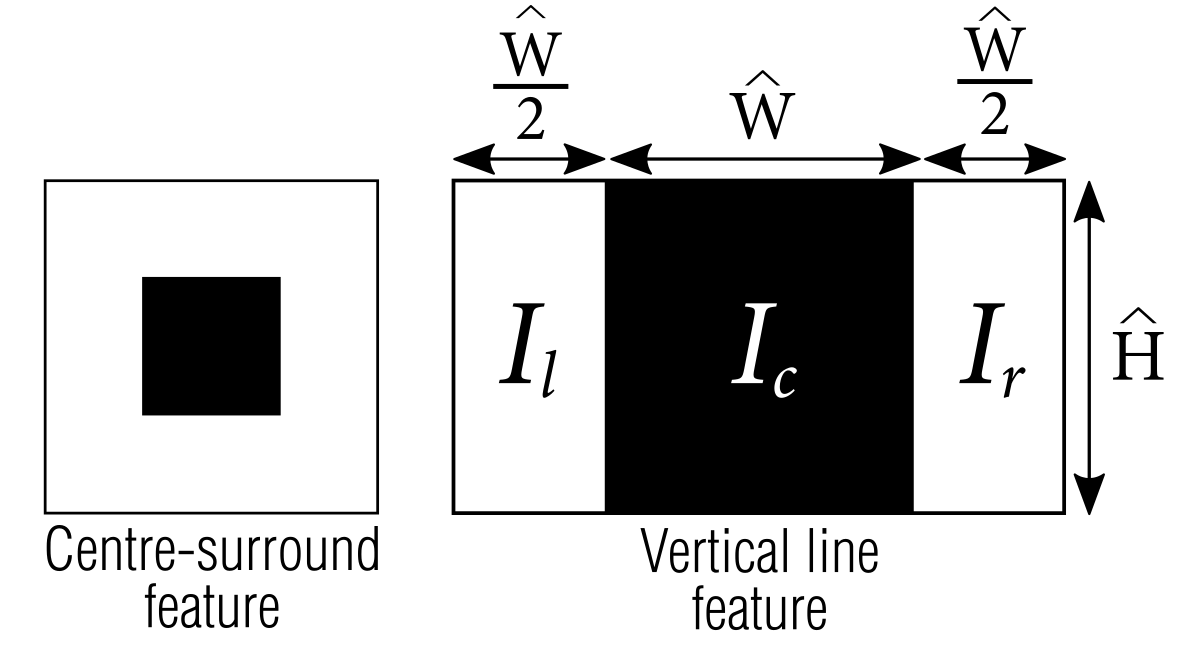}
		\caption{Two different Haar-like features. The vertical line feature is used in the pupil detection algorithm. Its dimensions are determined by pupil size predictions, $\hat{W}$ and $\hat{H}$. Three different values define the feature response, which are the summed pixel intensities of the left ($I_{l}$), centre ($I_{c}$) and right ($I_{r}$) areas.}
		\label{fig:haar_like_features}	
	\end{figure}

	An endless number of Haar-like features are available. In some other pupil detection algorithms  \cite{swirski2012} \cite{Kassner2014}, the centre-surround feature is used to obtain an approximate position of the pupil (see Figure \ref{fig:haar_like_features}), where it is assumed that the maximum response is obtained if the dark region overlaps with the pupil. Here, on the other hand, we opt for a vertical line feature instead, for two reasons. First, the primary feature in the image that might fool a basic feature detector are the eye lashes, because they can sometimes be as dark as the pupil. However, they are also quasi horizontally homogeneous in their brightness. So by looking at the contrast between the central area and the two areas it is horizontally flanked by, we can better discriminate between pupil and eye lashes. Second, the contrast of a vertical line feature is not as negatively affected by eye lashes when the eye is partially closed compared to the centre-surround feature, since eye lashes are generally located directly above or below the pupil, which is mostly ignored by the vertical line feature when it is centred on the pupil. 
	
	As mentioned before, the Haar-like feature detection method works by just looking at the difference in total intensity between the dark and light regions, disregarding the absolute intensity of either area. This does mean, however, that a relatively bright feature can potentially mislead the feature detector if it happens to be flanked by an even brighter area. For this reason, we consider both the intensity of the dark region as well as the contrast between dark and light regions when calculating the Haar-like feature response, $F_{H}$:
	\begin{equation}
	F_{H} =  -w_{1} \bar{i}_{c} + w_{2} (\frac{\bar{i}_{l} + \bar{i}_{r}}{2} - \bar{i}_{c})
	\label{eq:haar_feature_response}
	\end{equation}          
	
	We use the average pixel intensity $\bar{i}$ to calculate the response, instead of the summed intensity, because this makes the response invariant to the size of the fluctuating Haar-like feature area, which depends on the predicted width and height of the pupil. Both terms are modified by a weight constant $w$. We want to find the optimal combination of weights that leads to the greatest response for the pupil region and the smallest for non-pupil regions. From our data set, we obtain $\bar{i}$ values for the three regions at image locations that correspond with our measured pupil positions, but also at locations directly outside the pupil region (e.g. eye lashes). We then calculate $F_{H}$ for both groups and determine for which weight ratio the greatest separation between the two groups is found. This degree of separation is quantified by the test statistic of the two-sample Kolmogorov-Smirnov test, or \texttt{kstest2} in  MATLAB\textsuperscript{\textregistered} (R2016a, The MathWorks, Natick, MA, United States). Greatest separation is obtained by finding the maximum of the test statistic with respect to the unconstrained weight values, using the MATLAB function \texttt{fminsearch}. This process is performed on a random sub-set of the data. A different sub-set is used for evaluation. The optimal weight ratio is found for:
	\begin{equation*}
	\frac{w_{1}}{w_{2}} = 3.3
	\end{equation*}  
		
	One last confounding element that needs to be taken care of is the corneal reflection caused by the infra-red LED. This light can significantly disturb feature detection if it reflects off the cornea in front of the pupil, because it causes an increase in pupil brightness. We deal with this glint by first detecting it and then removing its influence if it overlaps with the centre rectangle of the vertical line feature. Glint detection is performed by convolving the AOI using the kernel given below in which every term is zero except in the centre and corners.  
	\begin{equation*}
	kernel = 
	\begin{pmatrix}
	-1 & 	   0 & \cdots & 	 0 & 	 -1 \\
	0 & \ddots &   	  &   	   & 	  0 \\
	\vdots &  		 & 	    1 &        & \vdots \\
	0 &  		 &  	  & \ddots &	  0	\\
	-1 & 	   0 & \cdots & 	 0 &	 -1 \\
	\end{pmatrix}
	\end{equation*} 
	
	A square kernel is used because the glint is often approximately circular. The width of this kernel, $W_{kernel}$, should be slightly larger than the diameter of the corneal reflection in the image. If the distance between IR-LED and eye varies little between individuals for the hardware set-up, which is generally the case, then this parameter can be set as a global constant.  The point of maximum response after convolution is then assumed to coincide with the glint position. This method provides robust detection as long as the glint overlaps with the pupil in the image, which is sufficient for our purposes because its position is irrelevant to us when it does not obscure the pupil. 
	
	\begin{figure}[ht]
		\centering
		\includegraphics[width=0.4\textwidth]{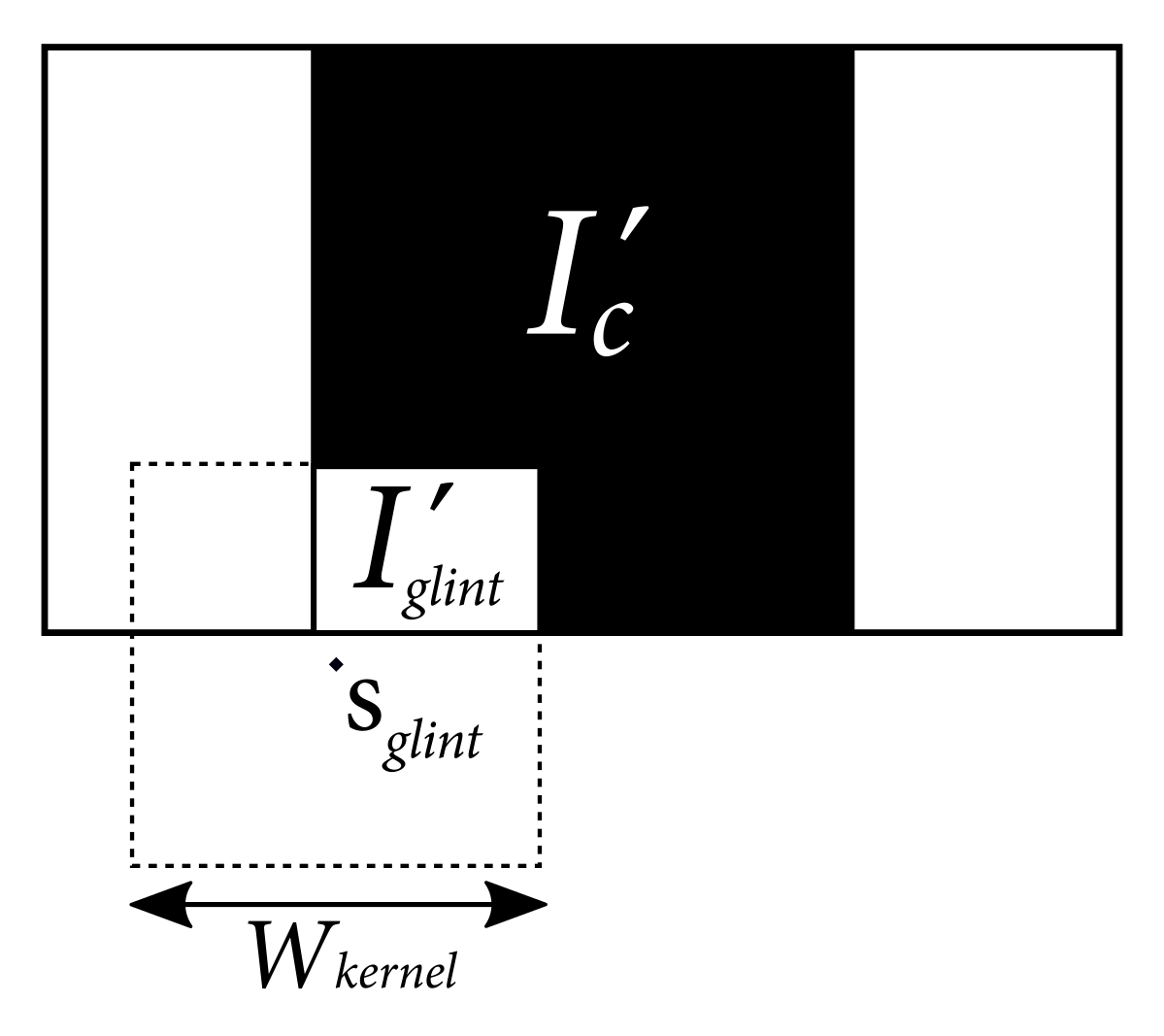}
		\caption{Removal of glint influence from Haar-like feature. The dashed rectangle indicates the position of the corneal reflection. The quantity $I'_{glint}$ is the summed intensity of the section of the glint area that overlaps with the central area of the Haar-like feature. The intensity $I'_{glint}$ is subtracted from the summed intensity of the central area to obtain the corrected intensity, $I'_{c}$.}
		\label{fig:haar_like_feature_glint}
	\end{figure}
	
	Once the glint has been detected, a square with the same size as the convolution kernel is centred on the glint position, $s_{glint}$. The summed intensity of the section of this square that overlaps with the dark area of the vertical line feature is subtracted from $I_{c}$ (see also Figure \ref{fig:haar_like_feature_glint}):
	\begin{equation}
	I'_{c} = I_{c} - I'_{glint}
	\end{equation}
	By doing the same for the surface areas, we can calculate a new average pixel intensity of the central region after glint removal:
	\begin{equation}
	\bar{i_{c}} = \frac{I'_{c}}{\hat{W} \times \hat{H} - A'_{glint}}
	\end{equation}
	\noindent
	Where $A'_{glint}$ is the surface area of the region that corresponds with $I'_{glint}$.
	
	The position $s_{H}$ of maximum Haar-like feature response can now be used to update $\hat{s}$. However, we must be cautious in using $s_{H}$, because it is more prone to errors than $\hat{s}$ when the certainty is high. Therefore, the greater the certainty, the less the new position prediction $\hat{s}_{new}$ will be based on $s_{H}$.
	\begin{equation}
	\hat{s}_{new} = s_{H} + c_{n}(\hat{s} - s_{H})
	\end{equation}
	Lastly, since these functions can potentially operate on the entire image (depending on the size of the AOI), it might be necessary to limit the number of iterations by down-sampling the image in order to achieve satisfactory computational speed. This should not appreciably affect accuracy of pupil detection, since we are only interested in an approximate position at this point. The original image resolution is immediately restored afterwards, before proceeding with the next processing steps. Further speed enhancement can be achieved during glint detection by only performing image convolution on pixels that have a brightness above a certain high threshold (e.g. $>200$ for 8-bit grayscale). 
	
	\FloatBarrier
	
	\subsection{Canny edge detection} 
	\label{sec:canny_edge_detection}
	
	After updating the approximate pupil position $\hat{s}$ in the previous step, Canny edge detection \cite{canny1986} is performed in an area with dimensions $W_{AOI} \times H_{AOI}$, centred around $\hat{s}$. The OpenCV \cite{opencv_library} implementation is chosen to carry out this task due to its computational efficiency. A Gaussian filter is applied beforehand. The Canny edge detector identifies and locates points of sharp changes in pixel intensity, which characterize boundaries of objects in the image \cite{maini2009}, and combines these points into thin line segments called edges. It transforms the AOI into a binary image, where all pixels that belong to an edge, also known as edge points, have been given a value of 1 and all other pixels a value of 0.  Here, an individual edge is defined as any collection of 8-connected edge points.
	
	\FloatBarrier
	
	\subsection{Morphological operation}
	
\begin{figure}[ht]
		\centering
		\includegraphics[width=0.5\textwidth]{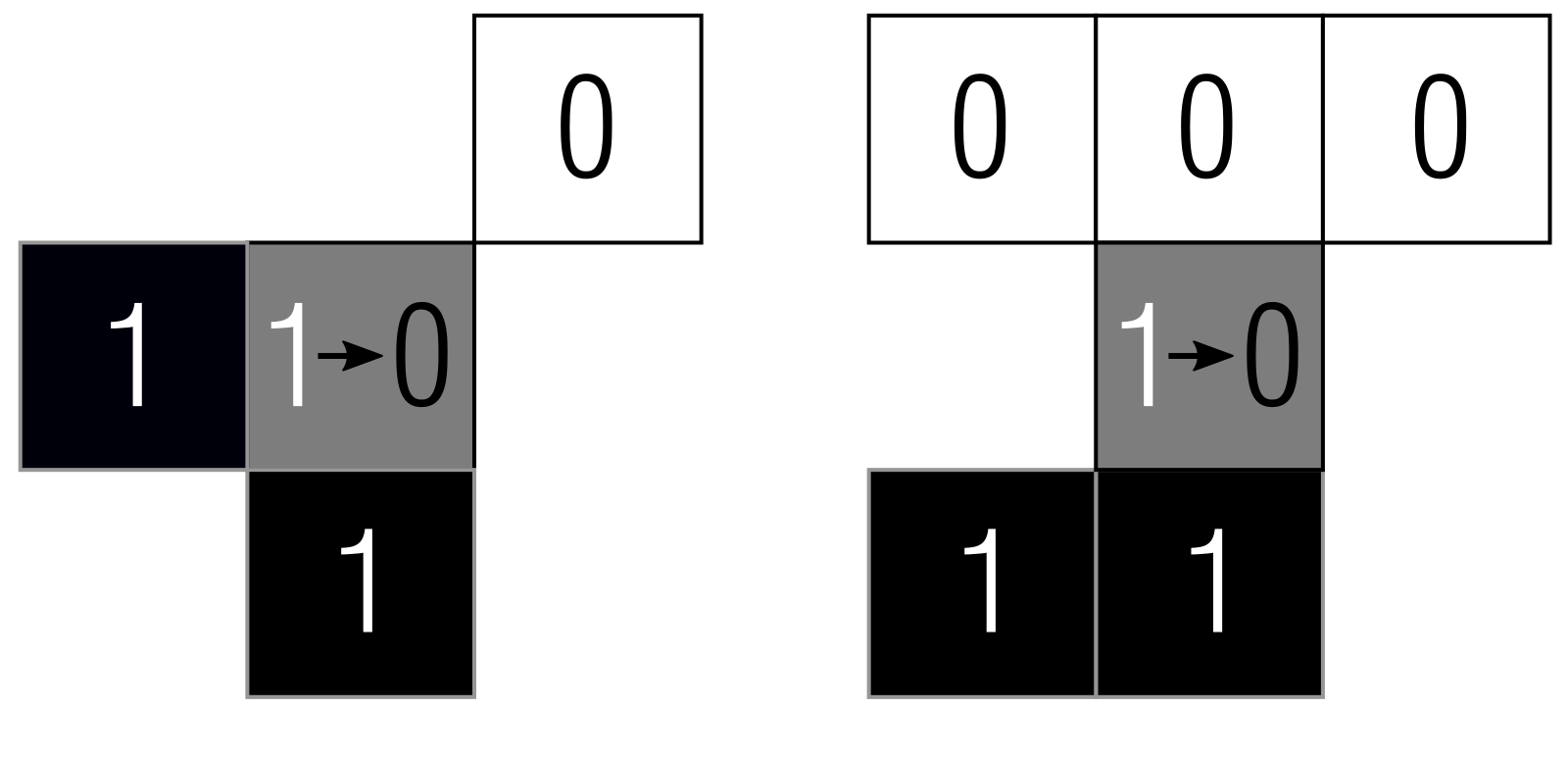}
		\caption{Edges are thinned by removing pixels from the edge if their 8-connected neighbourhood matches at least one of these two patterns. Black and grey pixels represent the pixels that belong to an edge. The patterns can be freely rotated and mirrored. The left pattern trims diagonally oriented edges, whereas the right pattern trims horizontally or vertically (after \SI{90}{\degree} rotation) oriented edges.}
		\label{fig:morph}	
	\end{figure}

	In the next part of the algorithm, a number of processing steps are performed on the detected edges, which attempt to filter out any edge that does not belong to the pupil-iris boundary. In order to speed up and simplify some of these processes, all edges need to be thinned to the minimum amount of pixels required to define it. During Canny edge detection, non-maximum suppression will have already significantly sharpened the edges, but not yet sufficiently for our intentions. So two morphological operations are applied to the image, which are illustrated in Figure \ref{fig:morph}, that trim the edges to single pixel thickness. Any edge points erased by this operation are not entirely discarded, but are given a special tag instead. Before fitting an ellipse on a sub-set of the detected edges (see section \ref{sec:ellipse_fitting}), these removed edge points are restored in order to obtain a more accurate fit. 
	
	\FloatBarrier
	
	\subsection{Edge selection}
	
	The first edge filter that is implemented is based on the fact that the pupil contour is likely to encircle or at least be close to the pupil position prediction $\hat{s}$. So we are going to select a sub-set of edges that are located at an acceptable distance from the pupil position estimate by sending out rays from $\hat{s}$ in eight directions, taking inspiration from the Starburst algorithm \cite{li2005}. The first edge that each ray encounters along its path is accepted, as well as any other edges that they come across after the first, provided they are within a radius:
	\begin{equation*}
	r = \frac{\Delta L}{2}
	\end{equation*} 
	
	By allowing the ray to continue beyond $r$ when no edge was encountered, we make the method more robust against pupil size predictions that are too small. Furthermore, by accepting all encountered edges within a radius $r$, and not just the first edge a ray runs into, we avoid detection failure when $\hat{s}$ happens to be (partially) enclosed by non-pupil edges (e.g. glint contour). Furthermore, very small edges are ignored for edge selection. This threshold is determined by the edge window length parameter $N_{l}$, which serves multiple purposes in the algorithm (see section \ref{sec:curvature_segmentation} for its value).
		
	\subsection{Edge classification}
	\label{sec:edge_classification}
	
	In the final step of our pupil detection algorithm, an ellipse needs to fitted on a combination of the previously selected edges. When multiple edges are available, it might be necessary to fit more than one ellipse and then choose the optimal one based on a few criteria described later (see section \ref{sec:ellipse_fitting}). We wish to avoid that scenario, however, because even though direct least squares fitting is relatively computationally inexpensive \cite{fitzgibbon1999} \cite{puatruaucean2012}, it can still be costly to fit many ellipses using this procedure in a single camera frame. Given $n$ edges, the maximum number of ellipses we would have to fit is equal to the total number of edge combinations:
	\begin{equation} \label{eq:combinations}
	\sum_{1 \leq k \leq n} \binom{n}{k} = 2^n - 1
	\end{equation}
	Which is the sum over all possible combination sizes $k$ chosen from $n$ (minus the null set). This value increases exponentially with respect to $n$, so it is important to keep the number of edges as small as possible. For this reason, a classification scheme is applied that divides edges into two categories: those that are part of the pupil-iris contour and those that are not. Edges that belong to the latter category are discarded. We base this classification on a number of different edge features, which are listed in Table \ref{table:edge_features}.
	
	One edge feature is the length $L$ of an edge, which is approximated by the sum of the distances between 8-connected neighbours. If the relative arrangement of two neighbouring edge points is in one of the four cardinal directions then the distance between them is 1, if it is in one of the four intercardinal directions then it is $\sqrt{2}$. This length measure will be sufficiently accurate due to the morphological operations that have been applied, which have left the edges with minimum thickness. 
	
	The variance of radius $\sigma_{r}$ tells us about the orientation of an edge. It will be close to zero if the edge encircles $\hat{s}$ at a distance that is roughly constant, which we expect to be the case for the pupil boundary. Other edges are likely to have different orientations, leading to larger $\sigma_{r}$ values. 
	
	The gradient $G_{r}$ here is different from the one encountered during glint detection. This gradient is calculated at each edge point by looking at the direction vector between the edge point and the position estimate $\hat{s}$. The difference in intensity between two pixels that lie along this direction on opposite sides of the edge point is the measure for $G_{r}$ at that particular position. The pixel value closer to $\hat{s}$ is subtracted from the one further away, so that a larger $G_{r}$ is obtained for gradients that transition from dark to light the further we move outwards. By considering the radial gradient, we add additional weight to edges belonging to the pupil boundary, because it marks the transition from dark to light in the radial direction, unlike many other edges. 
	
	The intensity $I$ is not simply given by the mean grayscale value of all edge points, but is instead calculated while taking into account the direction the edge curves in. Only pixels that lie on the inside of the edge curve are considered for $I$. Of course, for the pupil these pixels will be relatively dark, but for other edges this is not necessarily the case. How the location of the inside of the curve is determined relates to how the curvature $\kappa$ is calculated, which is described at a later stage (see section \ref{sec:curvature_segmentation}).
	
	\begin{table}[ht]
		\resizebox{\textwidth}{!}{%
			\begin{tabular}{ M{2cm} M{1cm} M{4cm} m{10cm} N}
				
				\textbf{Edge features} 
				& \textbf{Symbol} 
				& \textbf{Feature value, $F_{i}$} 
				& \textbf{Description} \\ 
				
				\hline
				Length 
				& $L$ 
				& $\displaystyle{\frac{\abs{ L - \hat{C} }}{\max(L,\hat{C})}}$ 
				& Summed distance between neighbouring edge points 
				& \\ [25pt]
				
				\hline
				Radius 
				& $r$ 
				& $\displaystyle{\frac{\abs{ r - \hat{C} / 2\pi }}{\max(r,\hat{C}/2\pi)}}$ 
				& Mean distance from each edge point to $\hat{s}$ 
				& \\ [25pt] 
				
				\hline
				Variance of radius
				& $\sigma_{r}$
				& $\displaystyle{\frac{\sigma_{r}}{\hat{C}}}$ 
				& Variance of the distance from each edge point to $\hat{s}$ 
				& \\ [25pt] 
				
				\hline
				Curvature 
				& $\kappa$ 
				& $\displaystyle{\abs{ \kappa - \hat{\kappa} }}$ 
				& Mean curvature of edge points 
				& \\ [25pt] 
				
				\hline
				Radial gradient 
				& $G_{r}$ 
				& $\displaystyle{\abs{ G_{r} - \hat{G}_{r} }}$ 
				& Mean gradient of edge points calculated in  radial direction from $\hat{s}$
				& \\ [25pt]
				
				\hline
				Intensity 
				& $I$ 
				& $\displaystyle{\abs{ I - \hat{I} }}$ 
				& Mean brightness of pixels on inside of edge curve
				& \\ [25pt]
				
		\end{tabular}}
		\caption{Edge features that the edge classification scheme is dependent on.}
		\label{table:edge_features}
	\end{table}
		
	From the available edge feature values, higher-level feature values $F_{i}$ are constructed via the expressions given in Table \ref{table:edge_features}, which are made invariant to the image size. For each feature value $F_{i}$, a score is calculated through a function that is unique to each feature. We distinguish the two classes using a linear combination of all scores $S_{tot}$, where the classification is controlled by a constant threshold value. The total score value is computed by:
	\begin{equation} \label{eq:score_function}
	S_{tot} = \sum_{i}^{6} w_{i} g_{i}(F_{i})
	\end{equation}
	Where $g_{i}$ is a Gaussian function with a certain set of parameters unique to feature $i$, and $w_{i}$ is a weight that signifies the importance of that particular feature in classification. The choice for a Gaussian function is made because it can easily be scaled in accordance with the frame-rate. The standard deviation of the function is increased with smaller frame-rates, because our predictions are likely to be less accurate, causing feature values of pupil boundary edges to fall farther from zero.  
	
\begin{figure}[ht]
		\centering
		\includegraphics[scale=1.10]{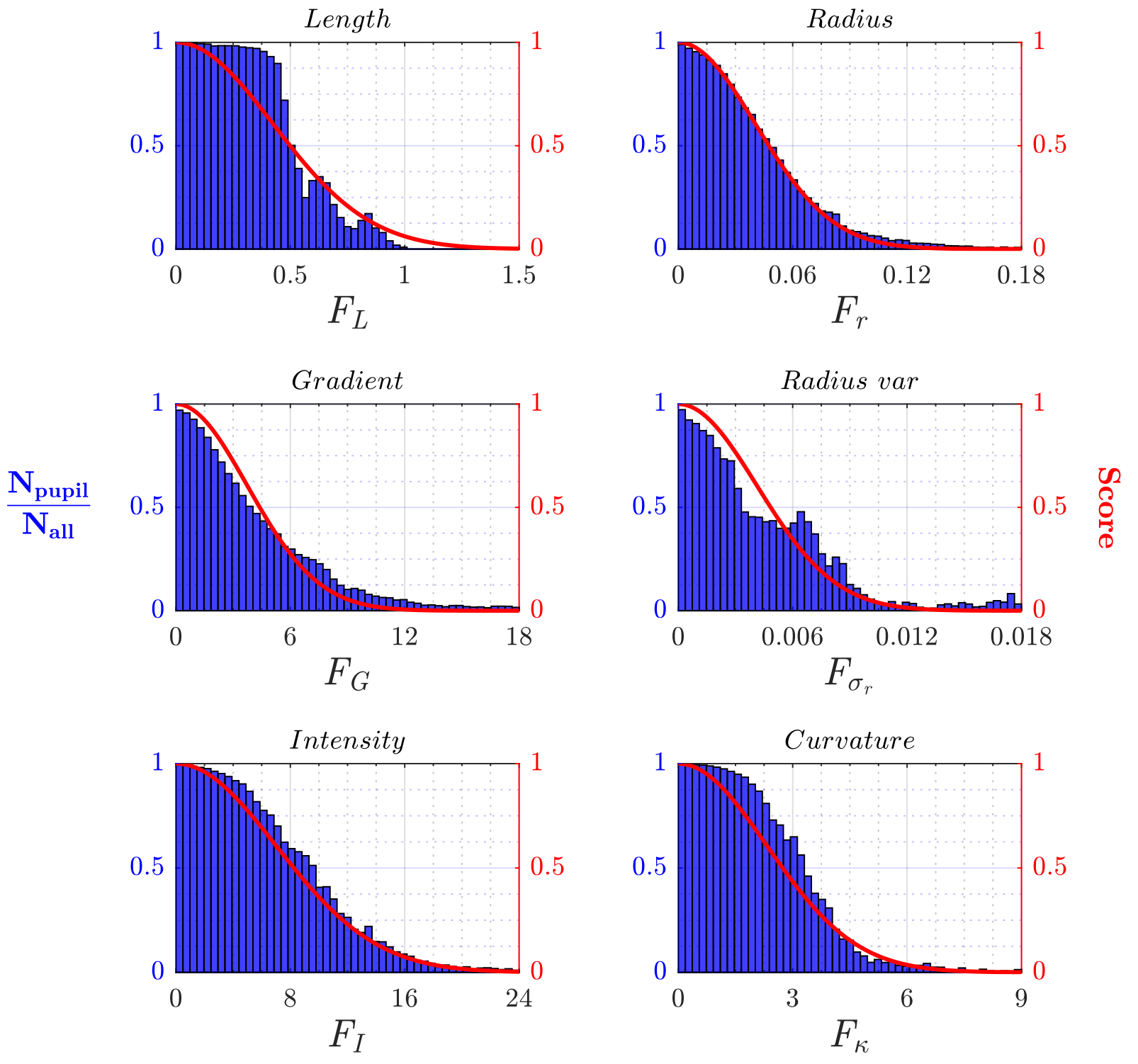}
		\caption{For each of the six feature values given in Table \ref{table:edge_features}, a histogram (blue bars) is plotted with the fraction of edges that are pupil edges in each bin. So, each bin gives the estimated probability that an edge with that particular feature value belongs to the pupil edge. The smaller the feature value, the more likely it is that the corresponding edge belongs to the pupil-iris contour. Gaussian functions have been fitted on the data (red curves), which convert the feature value to a score.}
		\label{fig:edge_gaussians}
	\end{figure}

	We are interested in determining which range of feature values we typically get for pupil-iris contour edges. This frequency of occurrence is then used to obtain the Gaussian functions needed for equation \ref{eq:score_function}. The frequency data for each feature value are extracted from one half of our data set and plotted in Figure \ref{fig:edge_gaussians}. To clarify, one half of the data (randomly selected) is used to find the Gaussian functions and weights, while the other half is set aside for testing the classifier. This labelled data set was acquired by placing edges in one of the two classes depending on whether they were part of the accepted ellipse fit or not. This classification works, because pupil boundary edges are significantly more likely to be fitted by our ellipse fitting method compared to non-pupil edges. In Figure \ref{fig:edge_gaussians}, the fitted Gaussian functions are plotted in red. Each fit was constraint by setting the function’s maximum to 1 and the position of the maximum to 0, which just leaves the standard deviation as a free parameter.	The distributions of feature values $F_{\kappa}$ and $F_{\sigma_{r}}$ have been limited to edges with $F_{L} \leq 0.75$, because the curvature and variance of radius are only relevant if the edge is reasonably long. Furthermore, when calculating the scores for $\kappa$ and $\sigma_{r}$ their weights are modified with $F_{L}$ according to:
	\begin{equation}
	w' = (1 - \beta F_{L})w
	\end{equation}
	Where $\beta$ is another type of weight factor, whose value is determined alongside the other weights. We must also consider how certain we are about our predictions when calculating the scores. Every weight is therefore multiplied with the relevant certainty value. A summary of how each weight is calculated is given below.
	\begin{align*}
	w'_{L} 			&= c_{A}w_{L} \\
	w'_{r} 			&= c_{S}w_{r} \\
	w'_{\sigma_{r}} &= c_{S}(1 - \beta F_{L})w_{\sigma_{r}} \\
	w'_{\kappa}		&= c_{A}(1 - \beta F_{L})w_{\kappa} \\ 
	w'_{G_{r}}      &= c_{S}c_{A}w_{G_{r}} \\
	w'_{I} 			&= c_{A}w_{I}
	\end{align*}
	
\begin{figure}[ht]
		\centering
		\includegraphics[width=0.8\textwidth]{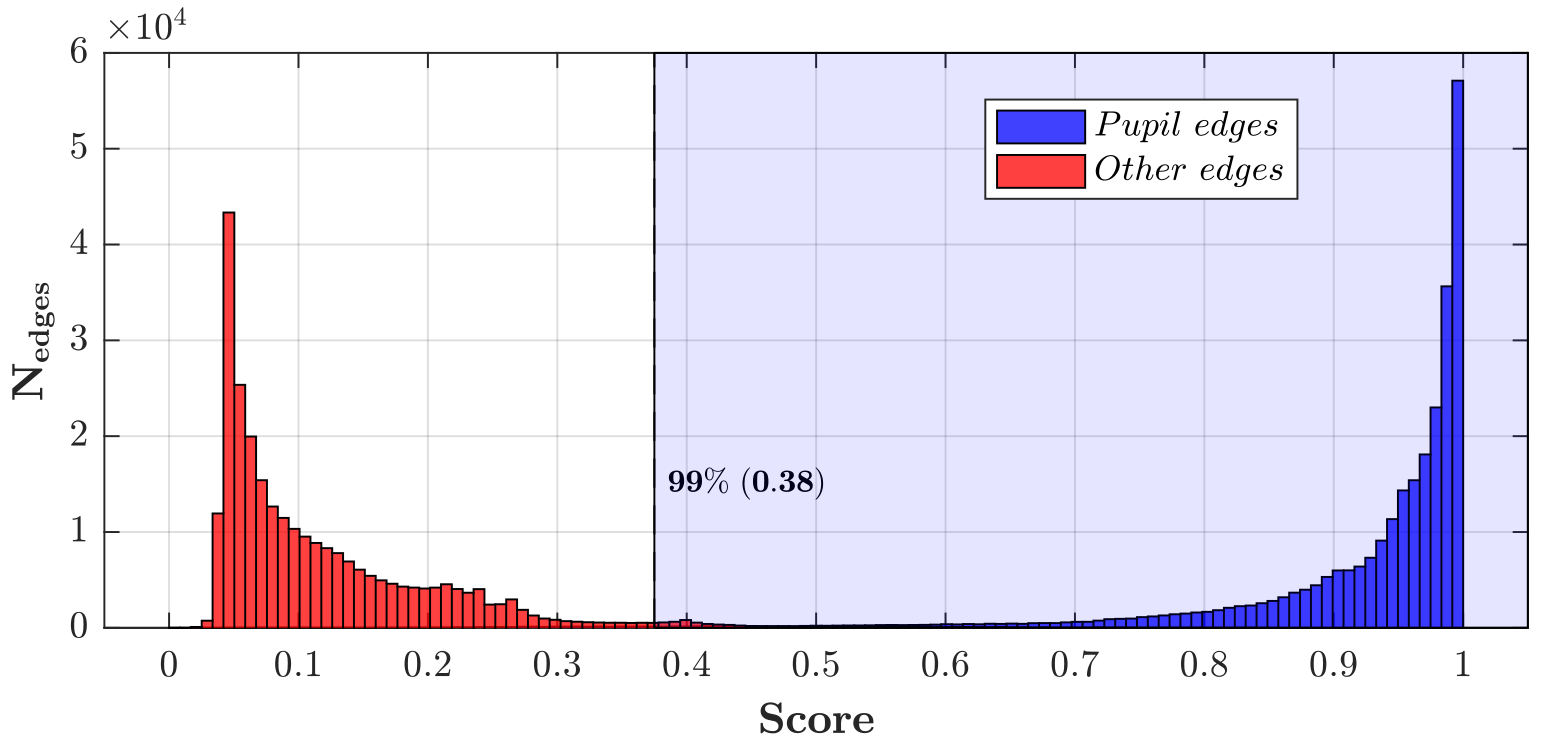}
		\caption{Score histogram of edges classified as either belonging to the pupil-iris contour (blue) or some other feature (red). Shown is the greatest separation between the two classes that was obtained with the weights given in Table \ref{table:weight_factors}. The vertical line denotes the threshold score for which 99\% of pupil edges are correctly classified.}
		\label{fig:edge_classifier}
	\end{figure}

	We can now train the classifier by assigning a weight to each feature value. To find the most optimal set of weights that best separates the two classes, we evaluate the degree of separation between them for a range of weight values. As in section \ref{sec:approximate_detection}, greatest separation between the two distributions is obtained by finding the maximum of the test statistic of the two-sample Kolmogorov-Smirnov test with respect to the weights. The weight configuration that corresponds with the greatest degree of separation is given in Table \ref{table:weight_factors}. The classifier is tested on the data that was not used to build the classifier. The resulting distributions are plotted in Figure \ref{fig:edge_classifier}, where the total score has been normalized to lie within 0 and 1. The vertical line marks the threshold position, $S_{\theta,edge}$, which corresponds with 99\% of pupil-iris contour edges being correctly classified ($S=0.38$). We make our threshold position dynamic with respect to the level of certainty, because we expect our score to be less accurate when the certainty becomes smaller. The actual threshold $S'_{\theta,edge}$ is then calculated by:
	\begin{equation} \label{eq:score_certainty}
	S'_{\theta,edge} = c_{S}c_{A}S_{\theta,edge}
	\end{equation}
	
\begin{figure}[ht]
		\centering
		\includegraphics[width=\textwidth]{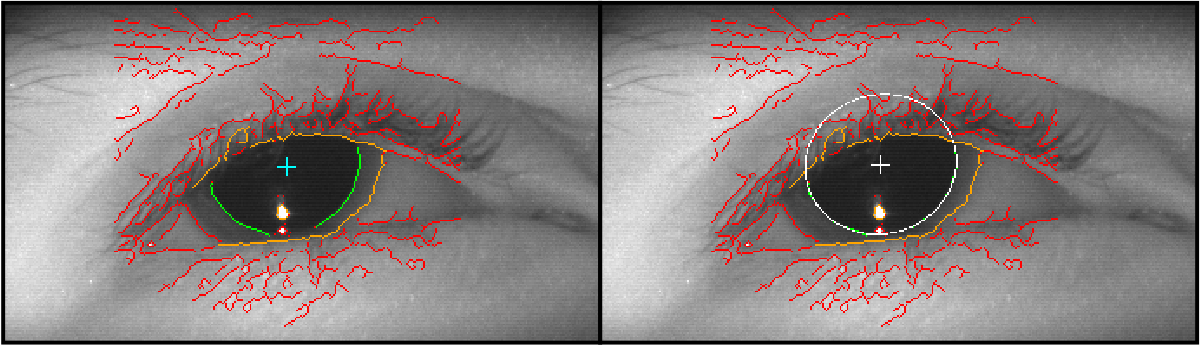}
		\caption{Edge classification example. Red edges have not been chosen by the edge selection method. Orange edges were chosen, but were filtered out due to edge classification. Green edges have been categorized as pupil-iris contour edges by our classifier and an ellipse is subsequently fitted on them (white outline, right image). Centre of ellipse is marked by the white cross. Teal cross indicates predicted pupil position.}
		\label{fig:eye_classification}
	\end{figure}

	In Figure \ref{fig:eye_classification}, we see the effect of the edge classification method on a sample image in which the eye is partially closed. The end result is that the pupil-iris contour edges are successfully extracted from the collection of Canny edges in the image, with the corresponding ellipse fit providing an accurate measure of the pupil position. 
	
	\begin{table}[ht]
		\centering
		\begin{tabular}{M{1.5cm} M{1.5cm} N}
			\textbf{Weight factor} & \textbf{Weight value} \\ 
			\hline
			$w_{L}$ 		
			& 0.7
			&\\[7pt] 
			\hline
			$w_{r}$ 		
			& 0.9 
			&\\[7pt]
			\hline
			$w_{\sigma_{r}}$ 
			& 1.2
			&\\[7pt] 
			\hline
			$w_{\kappa}$ 	
			& 1.4 
			&\\[7pt] 
			\hline
			$w_{G_{r}}$ 	
			& 0.7 
			&\\[7pt] 
			\hline
			$w_{I}$ 		
			& 1.4
			&\\[7pt] 
			\hline
			$\beta$			
			& 0.9
			&\\[7pt]
		\end{tabular}
		\caption{Optimal weight factors for maximum separation between the two edge distribution types.}
		\label{table:weight_factors}
	\end{table}
	
	\FloatBarrier
	
	\subsection{Edge segmentation}
	\label{sec:edge_segmentation}
	
	Owing to our classifier, we are now able to distinguish between edges that make up the pupil outline and edges that do not. However, this leaves one important class of edges not dealt with, namely those that only partially belong to the pupil boundary. Before fitting an ellipse on these types of edges, we have to split them first at the transition point between pupil edge and some other feature (e.g. eyelid), otherwise the accuracy of the ellipse fit will be adversely affected. For this purpose, three novel edge segmentation techniques have been developed, which are applied \textit{before} edge classification is carried out.
	
	\subsubsection{Path segmentation}
	\label{sec:path_segmentation}
	
	The first edge segmentation scheme is implemented based on the self-evident fact that the pupil is a non-intersecting closed shape. In other words, its perimeter is wholly described by a single non-branching path. So, if an edge consists of multiple branches, it definitely cannot exclusively be part of the pupil outline. Such edges are segmented by finding the non-branching path in the edge whose length comes closest to the circumference prediction $\hat{C}$ and separating it from the rest of the edge. To find this path, we will represent the edge as an undirected graph, which refers to the mathematical structure consisting of a collection of vertices (or nodes) joined by \textit{edges} with no orientation \cite[Chapter~6]{Newman2010}. To avoid confusion between the term edge here and how it was used before, we will use the term \textit{arc}, which is generally only used for directed graphs, to refer to connections between vertices in the network. The edge is converted to a graph through the following rules and definitions:
	
	\begin{enumerate}[noitemsep]
		\item	A branch vertex is any edge point that has three or more 8-neighbours
		\item	A terminal vertex is any edge point that has exactly one 8-neighbour
		\item	An arc is a collection of edge points that connects two vertices
		\item	Vertices that are 8-neighbours are combined into a single vertex
	\end{enumerate}
	
	After creating the graph, we use a recursive implementation of depth-first search to find every possible simple path, which is any path that does not traverse the same arc or vertex more than once, and then select the most optimal one. We locate all paths starting from the terminal vertices first and keep track of the vertices we already started with so we do not end with them. This prevents acyclic path repetitions, because for our purposes a specific path between two vertices is the same in the reverse direction.
	
	A caveat should be added. The task of finding all possible simple paths in an undirected graph is NP-hard, since the more basic \textit{longest path problem} is already NP-hard \cite{Cormen2001}. However, we are dealing with relatively tiny networks here with usually only a few connections, which allows for this problem to be quickly solved.
	
	\begin{figure}[ht]
		\centering
		\includegraphics[scale=1.40]{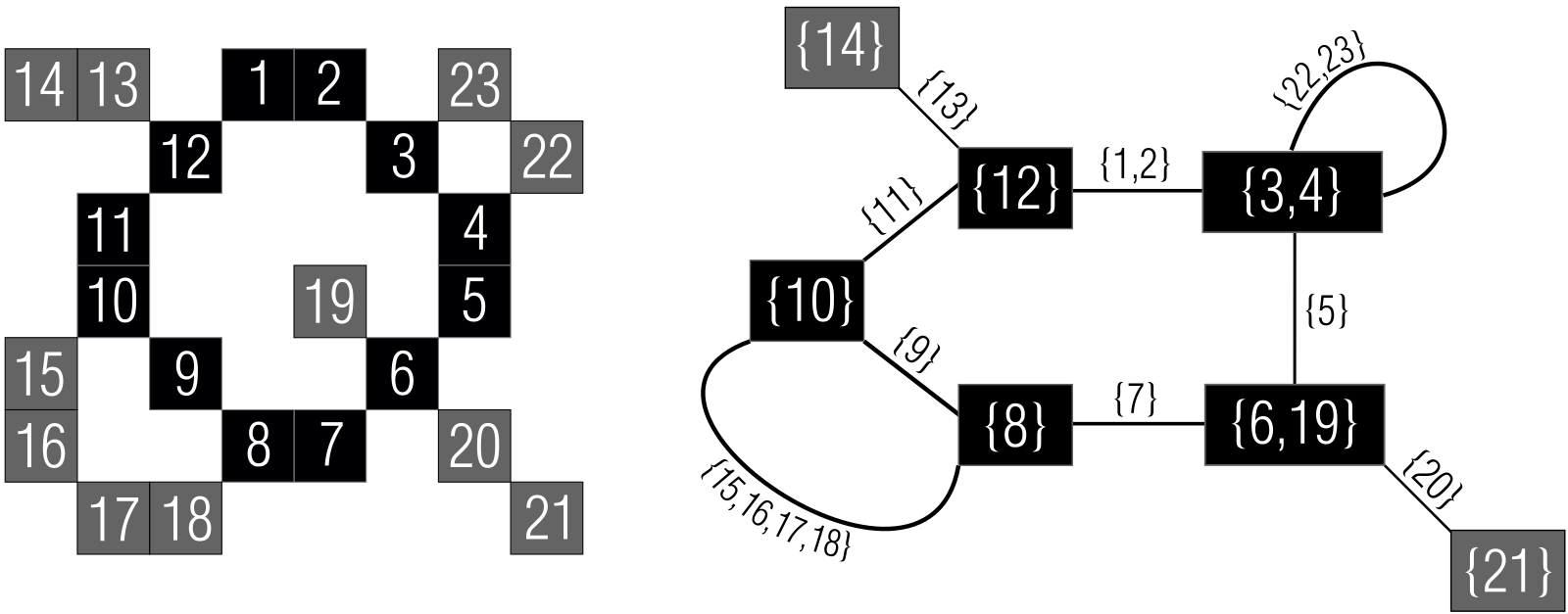}
		\caption{\textit{(Left)} Pixel representation of a sample Canny edge, where each pixel has been given its own unique label. \textit{(Right)} Graph representation of the same edge. The pixels contained by the vertices (boxes) and the arcs (lines) are given in set notation. Black pixels or vertices indicate the desired path.}
		\label{fig:edge_graph}
	\end{figure}
		
	In Figure \ref{fig:edge_graph}, a sample edge is depicted along with its graph representation. The structure of this graph is an exaggerated version of what one typically encounters. The total number of possible simple paths in this graph is equal to 139, which is still a manageable quantity, even when having to process multiple edges. However, this number can be brought down further, while also dealing with noise. We remove arcs from the edge if they contain fewer edge points than the edge window length $N_{l}$ when they are attached to a terminal vertex, are a self-loop or when they connect two vertices that are already connected by an arc, in which case the shortest of the two arcs is kept. As an example, when these filters are applied to the edge in Figure \ref{fig:edge_graph}, all grey edge pixels are removed. 
	
	Generally, several paths will be detected, one of which has to be accepted. Our decision is made based on a few criteria. Cyclic paths always take precedence over acyclic paths, since the ideal pupil outline is cyclic. However, if the pupil boundary is described by a cycle then it has to be wholly represented by that cycle, so the length $L$ of such a cyclic path has to match the expected circumference of the pupil boundary, where the acceptable range is given by: 
	\begin{equation*}
	C_{min} \leq \hat{C}(1 - \delta_{\theta,C}) \leq L \leq \hat{C}(1 + \delta_{\theta,C}) \leq C_{max}
	\end{equation*}
	
	The lower limit of this criterion must not apply for acyclic paths, because multiple acyclic paths can make up the pupil boundary. We could enforce the upper limit, though, since the length of an acyclic path should not be greater than $C_{max}$ either, but this is rejected in favour of a different solution (see section \ref{sec:length_segmentation}). Finally, any arcs that were not filtered out, but also not included in the final accepted path are not discarded. Instead, new graphs are made using these remaining edges and the process is repeated.
	
	\FloatBarrier
	
	\subsubsection{Curvature segmentation}
	\label{sec:curvature_segmentation}
	
	Another property of the pupil-iris contour is its smooth curvature. Abrupt changes in curvature along its path are most likely caused by obstructions of the pupil periphery by the eyelid, eye lashes or corneal reflection from the IR-LED. Zhu et al. \cite{zhu1999} developed an algorithm that exploits the curvature characteristic of the pupil boundary in order to separate it from these types of artefacts. The location of an occlusion or \textit{breakpoint} in the pupil boundary is determined by checking if the curvature $\kappa$ at a certain point on the perimeter is above an upper threshold  ($\kappa_{max}$) or below a lower threshold ($\kappa_{min}$). The pupil boundary is then segmented according to a number of heuristics that use the distances between detected breakpoints to detect different types of occlusions.
	
\begin{figure}[ht]
		\centering
		\includegraphics[scale=0.70]{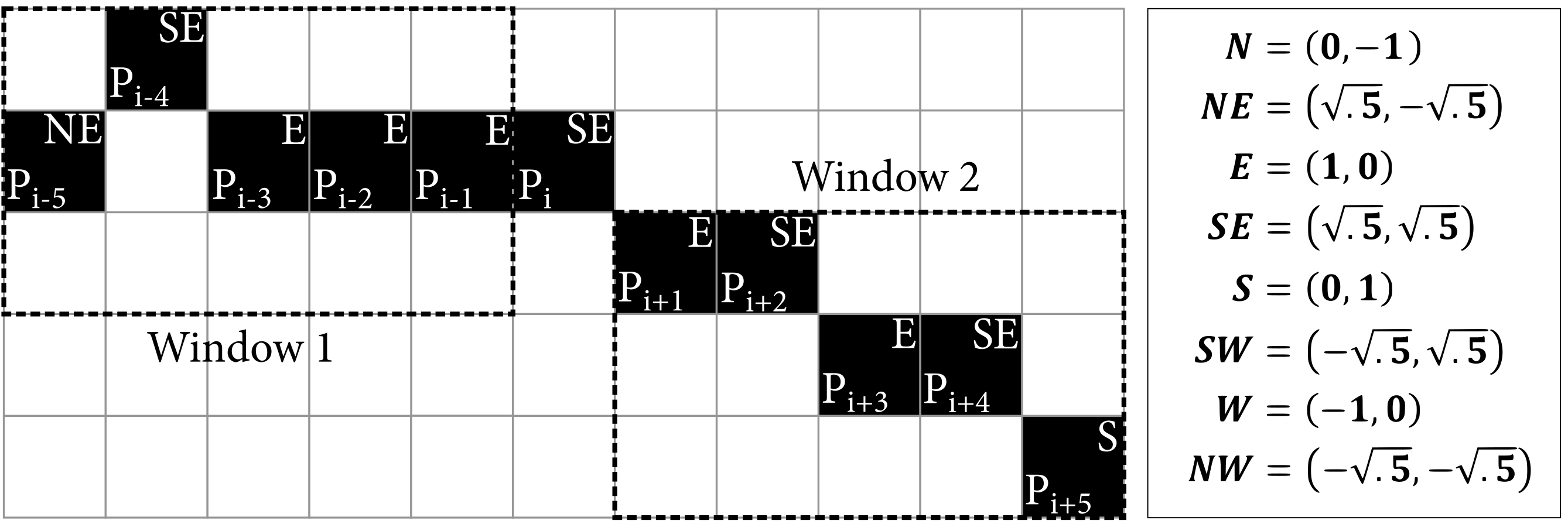}
		\caption{Example of curvature calculation procedure. Each pixel in the edge is given a (cardinal) direction label, which corresponds with one of the vectors displayed on the right. Curvature is calculated at pixel $P_{i}$ by finding the angle between the vector tangents of the two windows on either side of it.}
		\label{fig:edge_curvature}
	\end{figure}

	Inspired by this approach, we have developed our own segmentation algorithm based on edge curvature. The method that Zhu et al. used to calculate the curvature is repeated here, but with a few slight alterations. One change is made in the way direction coding is performed, because we are interested in calculating the curvature of Canny edges and not the boundary of a ‘blob’ of pixels that is generated after application of a brightness threshold. The method works by scanning the 8-connected environment of each edge point and assigning a vector label to the edge point depending on the relative position of its neighbour. For example, starting from the central edge point $P_{i}$ in Figure \ref{fig:edge_curvature}, we scan its 8-neighbours and locate $P_{i+1}$ in the south-east direction, which corresponds with a vector of ($\sqrt{.5}$,$\sqrt{.5}$) or a cardinal direction label of SE. We then do the same for $P_{i+1}$ and so forth, until every edge point has been given a label.
	
	After obtaining all direction labels, the curvature $\kappa$ at each edge point is calculated. Theoretically, the (signed) curvature at a specific point on a curve is defined \cite{patrikalakis2009} as the rate of change of the tangential angle $\phi$ with respect to the arc length $s$. This exact definition of $\kappa$ is approximated by placing two windows on either side of the edge point we are calculating $\kappa$ for and finding the mean unit vector for each window, where the size of a window is equal to the edge window length $N_{l}$. The difference in tangential angles $\Delta \phi$ between the two mean vectors, divided by $N_{l}$, is then the signed curvature at that point:
	\begin{equation*}
	\kappa = \frac{d\phi}{ds} \approx \frac{\Delta \phi}{N_{l}}
	\end{equation*}
	
	When determining the curvature for $P_{i}$ in Figure \ref{fig:edge_curvature}, we calculate the first vector tangent $\boldsymbol{\vec{T}_1}$ for edge points running from $P_{i-5}$ to $P_{i-1}$ (i.e. $N_{l} = 5$ in this example) and the second $\boldsymbol{\vec{T}_2}$ for points from $P_{i+1}$ to $P_{i+5}$. The angle difference $\Delta \phi$ is then given by:
	\begin{equation*}
	\Delta \phi = \atantwo (T_{2,y},T_{2,x}) - \atantwo (T_{1,y},T_{1,x})
	\text{  with  }   
	\Delta \phi = 
	\begin{cases}
	\Delta \phi - 2\pi  & \text{if } \Delta \phi >  \pi\\
	\Delta \phi + 2\pi  & \text{if } \Delta \phi < -\pi\\
	\end{cases}
	\end{equation*}
	
	For edges belonging to the pupil boundary, we expect that each edge point has the same curvature sign, since the boundary is elliptical. However, whether the majority sign is negative or positive depends on the scanning direction, which we do not control here. In order to properly set a lower curvature threshold it is crucial that each edge has the same majority curvature sign. For this reason, we count the number of positive and negative curvatures in the edge. If there are more negative curvatures than positive ones, all signs are inverted.
	
	\noindent
	We also determine the vector difference $\Delta\boldsymbol{\vec{T}_1}$ between $\boldsymbol{\vec{T}_1}$ and $\boldsymbol{\vec{T}_2}$:
	\begin{equation*}
	\Delta\boldsymbol{\vec{T}_1} = \boldsymbol{\vec{T}_2} - \boldsymbol{\vec{T}_1}
	\end{equation*}
	
	\noindent
	The vector $\Delta\boldsymbol{\vec{T}_1}$ points towards the centre of curvature. This direction is used when calculating the edge intensity in edge classification. 
	
	Now that we have a measure of the curvature for every edge point, we can use it to segment edges at the intersection between distinct features in the image. To do this, we do not use any of the heuristics developed by Zhu et al. Instead, our curvature segmentation algorithm works on the basis of only one simple rule: an edge is segmented at every breakpoint. This makes it far less specific and requires fewer parameters. The only parameters we need to specify are the values for $\kappa_{min}$ and $\kappa_{max}$. Like Zhu et al., we could set these to a constant value for a given pupil, with $\kappa_{min} = -\kappa_{max}$. However, this seems like a poor decision for two reasons. First of all, this heavily overestimates the expected lower curvature limit. Since the pupil contour is elliptical, the signed curvature should theoretically not become negative, as mentioned before. This means that $\kappa_{min}$ should be much closer to zero. Second of all, the expected values for $\kappa$ are very dependent on where the eye is looking. This is because the range of $\kappa$ is not only determined by the size of the pupil, but also on its shape. The circumference $C$ of the pupil is inversely proportional to the mean of $\kappa$. The aspect ratio $AR$, on the other hand, is inversely proportional to the size of the range of $\kappa$. When the eye is looking straight ahead, the pupil can be reasonably approximated by a circle (i.e. $AR = 1$), which means that $\kappa$ will be constant around the perimeter, so its range is minimal. However, when the eye is looking up, the pupil shape is more eccentric, so $AR$ will be smaller. This causes the curvature at the antipodal points of the semi-minor axis to decrease and the curvature at the semi-major axis antipodes to increase, broadening the range of $\kappa$.
	
	\begin{figure}[ht]
		\centering
		\includegraphics[width=\textwidth]{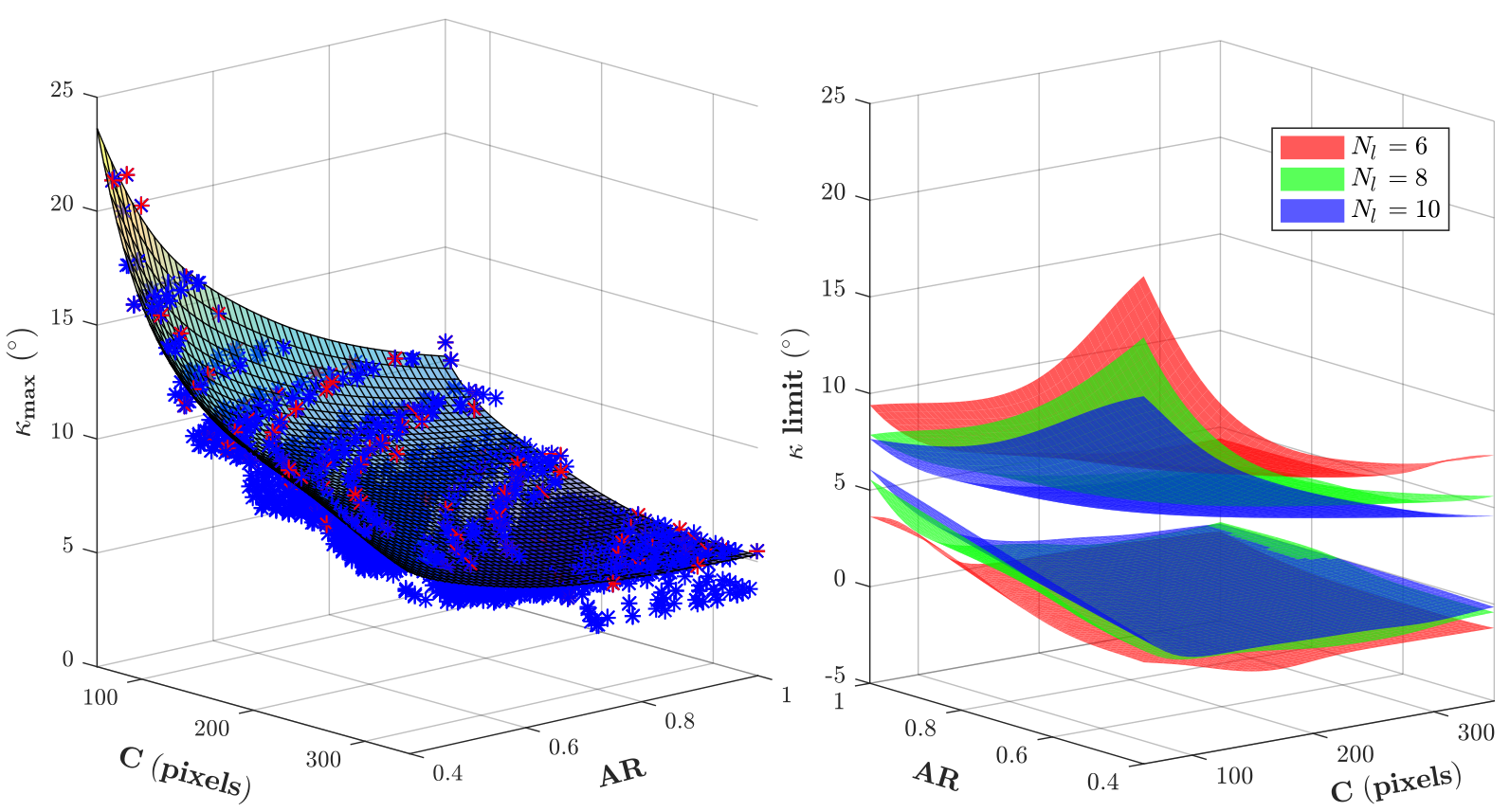}
		\caption{\textit{(Left)} Neural network fit of maximum edge curvature as a function of pupil circumference and aspect ratio for an edge window length of 5. The function is represented by the mesh surface, which is fitted on the red data points, indicating local maxima. \textit{(Right)} Dependence of minimum and maximum edge curvature on circumference, aspect ratio and edge window length, $N_{l}$. Bottom three surfaces are for $\kappa_{min}$ and top three for $\kappa_{max}$.}
		\label{fig:curvature_3D_3}
	\end{figure}
	
	To ensure that $\kappa_{min}$ and $\kappa_{max}$ depend on the size and shape of the pupil, these parameters are turned into dynamic thresholds that are automatically updated according to our predicted values $\hat{C}$ and $\hat{AR}$. The thresholds are based on the lower and upper limit of the $\kappa$ range we would expect to get for an ellipse with a circumference and aspect ratio equal to $\hat{C}$ and $\hat{AR}$. To find the relation between the range of $\kappa$ with respect to $C$ and $AR$, we measured the minimum and maximum $\kappa$ values of 3600 artificial pupils, which are solid black ellipses on a white background. These ellipses varied in circumference between 30 and 380 pixels, and in aspect ratio between 0.15 and 1.00. Furthermore, to investigate how the edge curvature depends on the curvature window size, the $\kappa$ range of each ellipse was evaluated multiple times in each frame for different $N_{l}$ values between 5 and 11. After obtaining all $\kappa_{min}$ and $\kappa_{max}$ data points, neural networks are trained to fit functions that take $C$ and $AR$ as their input and give $\kappa_{min}$ or $\kappa_{max}$ as their output. This task is performed by the MATLAB routine \texttt{fitnet} using Levenberg-Marquardt with 5 hidden nodes. The functions are only fitted on the largest $\kappa_{max}$ values or on the smallest $\kappa_{min}$ values that were measured in a specific circumference and aspect ratio bin. This is to ensure that we get the true curvature limit for the $C$ and $AR$ pair. A sample fit for $\kappa_{max}$ is displayed on the left side of Figure \ref{fig:curvature_3D_3}. The dependence of $\kappa_{min}$ and $\kappa_{max}$ on $N_{l}$ is shown on the right side of Figure \ref{fig:curvature_3D_3}. We can observe that the $\kappa$ range narrows with increase in $N_{l}$. This is because the greater the window size, the less refined our measurement is. We lose information about local curvature maxima or minima, causing them to be somewhat averaged out, which leads to a smaller $\kappa$ range. In this sense, we want to have the smallest possible value for $N_{l}$, but larger $N_{l}$ values make our $\kappa$ measurement more robust against noise. The default value for $N_{l}$ is set to 7, which was found to give good results for our set-up. However, it can be desirable to turn $N_{l}$ into a dynamic parameter that is reduced when $\hat{C}$ becomes smaller, because the smaller the pupil is, the more its contour edge will be affected by tiny interferences, which may go unnoticed when $N_{l}$ is set too high.
	
	\FloatBarrier
	
	The threshold values obtained here are curvature limits for ideal ellipses. In practice, these thresholds need to be offset (up and down) by a few degrees to account for inaccuracies in our predictions and for natural deviation of the pupil from the ellipse shape. The final thresholds are therefore determined by:
	\begin{align}
	\kappa_{max} &= h(\hat{C}(1 - \delta_{\theta,C}), \hat{AR} - \delta_{\theta,AR}) + offset \\
	\kappa_{min} &= h(\hat{C}(1 - \delta_{\theta,C}), \hat{AR} - \delta_{\theta,AR}) - offset
	\end{align}
	
	\noindent
	Where $h(C,AR)$ is the fitted function. The variable $\delta_{\theta,AR}$ is calculated in a similar way to $\delta_{\theta,C}$ (see equation \ref{eq:delta_circumference}):
	\begin{equation} \label{eq:delta_aspect_ratio}
	\delta_{\theta,AR} = (1 - c_{AR}) (1.0 - \delta''_{\theta,AR}) + \delta''_{\theta,AR} \\
	\end{equation}	
	
\begin{figure}[ht]
		\centering
		\includegraphics[scale=0.80]{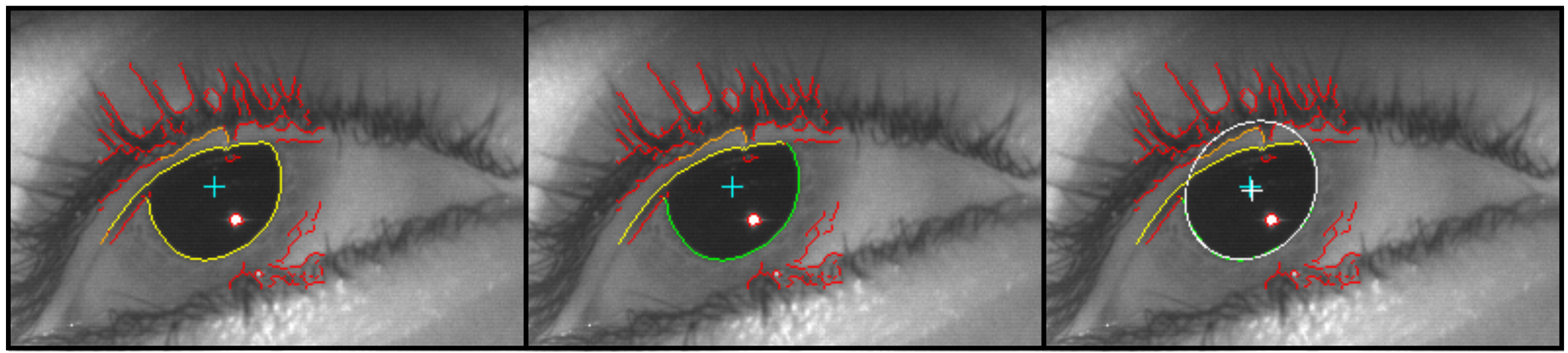}
		\caption{Curvature segmentation example. \textit{(Left)} Yellow edge is part of pupil boundary as well as eyelid. \textit{(Centre)} Edge is segmented at transition point between both features, resulting in new pupil edge section (green). \textit{(Right)} Ellipse (white outline) is fitted on segmented pupil edge. Centre of ellipse is marked by the white cross. Teal cross indicates predicted pupil position.}
		\label{fig:eye_curvature}
	\end{figure}

	Figure \ref{fig:eye_curvature} shows the effect of curvature edge segmentation on a sample image in which the eye is partially closed. The pupil-iris outline is detected by Canny edge detection as one continuous edge that is also part of the upper eyelid. Through curvature segmentation we are able to split the edge up into two sections corresponding to the distinct features, after which the pupil edge can be successfully identified with edge classification.
	
	\FloatBarrier
	
	\subsubsection{Length segmentation}
	\label{sec:length_segmentation}
	
\begin{figure}[ht]
		\centering
		\includegraphics[scale=1.00]{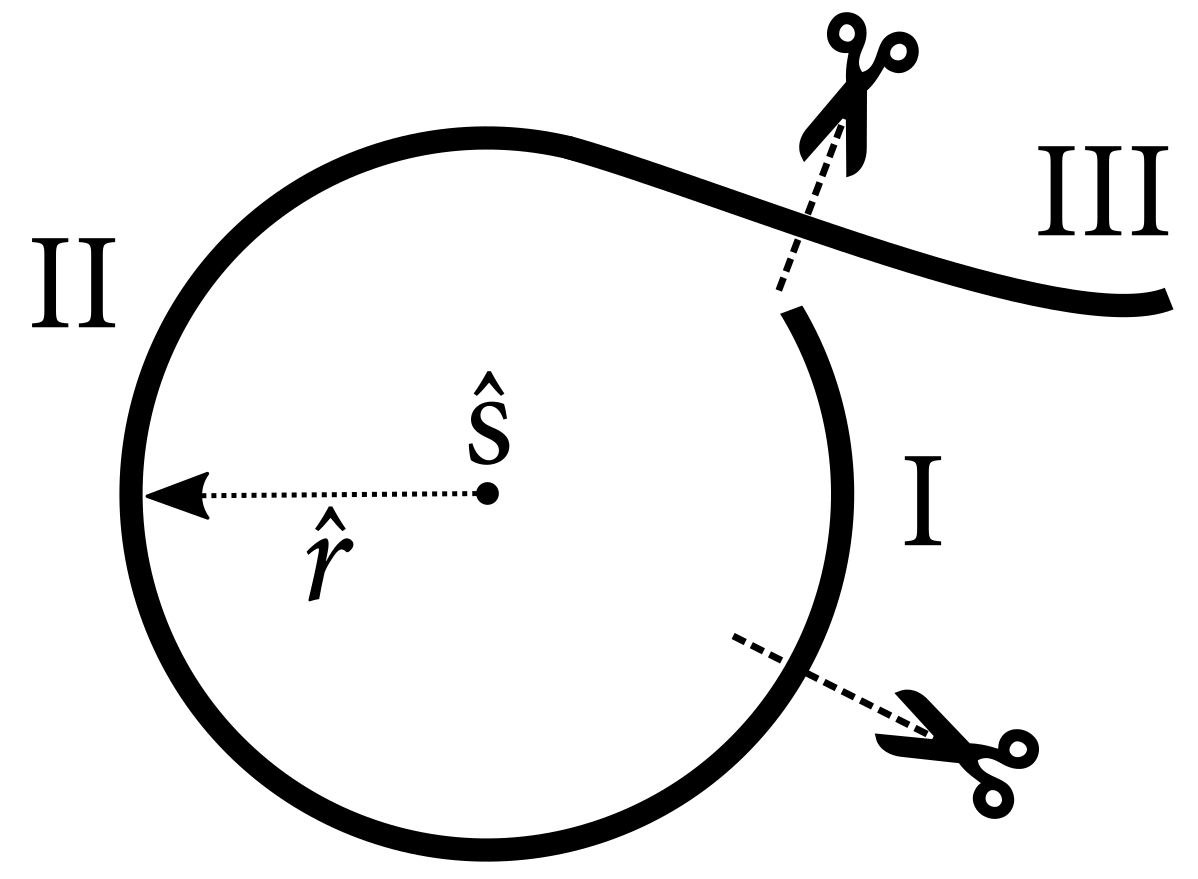}
		\caption{Length segmentation procedure. See main text for further details.}
		\label{fig:length_segmentation}
	\end{figure}

	If the pupil circumference prediction in a given frame is accurate, then no edge that entirely resides on the pupil boundary should be longer than this prediction.  When the edge length $L$ is greater than $\hat{C}$, we segment the edge so that one of the two parts has a length that is approximately equal to $\hat{C}$. The question, however, is where to make the separation. We make this decision by temporarily dividing the edge into three sections, which is graphically shown in Figure \ref{fig:length_segmentation}. The edge is cut in such a way that the length of the edge body (\textit{II}) plus the length of either of the two edge tails (\textit{I} or \textit{III}) is equal to $\hat{C}$, where $\hat{C} = 2\pi\hat{r}$ in the figure. 
	
	We now investigate which of the two edge tails resembles the body the most with respect to its features. The closest match is re-attached to the central section, while the other one is left segmented. The level of similarity is quantified using the score functions we developed for edge classification, but instead of using the difference between an edge feature value and the corresponding predicted value to calculate the score, we will work with the difference between feature values of the tail compared to the body. The expressions for calculating $F_{i}$ and $w'_{i}$ are given in Table \ref{table:length_weights}. The length weight is set to zero, since $F_{L}$ is equal for both edge tails. We only use $F_{L}$ to modify the weights of $\sigma_{r}$ and $\kappa$. The certainty terms $c_{A}$ are set to 1 in the weight expressions, because we are not using any predicted values to calculate $F_{i}$, so their accuracy is irrelevant here. However, the accuracy of $r$, $\sigma_{r}$ and $G_{r}$ does depend on $c_{S}$, hence it remains. The scores for each tail are calculated using equation \ref{eq:score_function}, and the edge tail with the highest score is reconnected to the edge body. In Figure \ref{fig:length_segmentation}, we expect edge \textit{I} to have the highest score, since its $r$, $\kappa$ and $\sigma_{r}$ are very close to edge \textit{II}, so edge \textit{III} is left severed. 
	
	\begin{table}[ht]
		\centering
		\begin{tabular}{ M{2cm} M{4cm} M{4cm} N}
			
			\textbf{Edge features} & \textbf{Feature value, $F_{i}$} & \textbf{Weight, $w'_{i}$} \\ 
			
			\hline
			Length 
			& $\displaystyle{\frac{\abs{ L_{1} - \hat{C} }}{\max(L_{1},\hat{C})}}$ 
			& 0 
			& \\ [25pt]
			
			\hline
			Radius 
			& $\displaystyle{\frac{\abs{ r_{1} - r_{2} }}{\max(r_{1},r_{2})}}$ 
			& $c_{S}w_{r}$ 
			& \\ [25pt]
			
			\hline
			Variance of radius 
			& $\displaystyle{\frac{\abs{\sigma_{r,1} - \sigma_{r,2}}}{\hat{C}}}$ 
			& $c_{S}(1 - \beta F_{L})w_{\sigma_{r}}$ 
			& \\ [25pt]
			
			\hline
			Curvature 
			& $\displaystyle{\abs{ \kappa_{1} - \kappa_{2} }}$ 
			& $(1 - \beta F_{L})w_{\kappa}$ 
			& \\ [25pt]
			
			\hline
			Radial gradient 
			& $\displaystyle{\abs{ G_{r,1} - G_{r,2} }}$ 
			& $c_{S}w_{G_{r}}$ 
			& \\ [25pt]
			
			\hline
			Intensity 
			& $\displaystyle{\abs{ I_{1} - I_{2} }}$ 
			& $w_{I}$ 
			& \\ [25pt]
			
		\end{tabular}
		\caption{Edge features that are taken into account for length segmentation. Included are the expressions for the corresponding feature value and weight factor.}
		\label{table:length_weights}
	\end{table}
	
	\FloatBarrier
		
	\subsection{Ellipse fitting}
	\label{sec:ellipse_fitting}
	
	Having segmented and classified the edges obtained with Canny edge detection, we can now fit an ellipse on one or more of them, using one of several possible methods. Search- and voting-based schemes, such as Hough transform and Random Sample Consensus (RANSAC), are often implemented in pupil detection algorithms because they are robust to outliers, however they are also computationally expensive \cite{Fuhl2015}\cite{raguram2008}\cite{swirski2012}. Since we have performed numerous segmentation routines to remove any potential outliers, we instead choose the faster, but more sensitive, direct least squares fitting method \cite{fitzgibbon1999}. Owing to its computational efficiency, we are able to fit an ellipse multiple times in the same frame and then choose the most optimal one. However, recall equation \ref{eq:combinations}. We cannot have a large set of remaining edges, otherwise the number of possible edge combinations to fit an ellipse on will be overwhelming. So we choose up to a fixed number of available edges classified as lying on the pupil-iris contour, favouring edges with a higher score. The maximum number that we choose is set to 4 by default, which translates to 15 possible combinations.
	
	Fitting that many ellipses in one frame is still too demanding, but we are going to reduce their numbers further by requiring that the expanse of each edge combination must reasonably correspond with our width and height predictions of the pupil. This also means that we require that a significant portion of the pupil-iris contour is visible in the image and has been detected before fitting an ellipse, because the fit will be gravely inaccurate otherwise.  We calculate the range of the edge combination by finding the minimum and maximum $x$ and $y$-positions in the collection of edge points and set the following criterion: 
	\begin{alignat*}{2}
	0.3(\hat{W} - \Delta l) &\leq x_{max} - x_{min} &&\leq \hat{W} + \Delta l \\
	0.3(\hat{H} - \Delta l) &\leq y_{max} - y_{min} &&\leq \hat{H} + \Delta l
	\end{alignat*}
	\noindent
	Where: 
	\begin{equation}
	\Delta l = \frac{\hat{C}\delta_{\theta,C}}{\pi}
	\end{equation}
	
	These thresholds ensure that cases where the vast majority of the pupil is obscured by the eyelid are ignored. A limit is also imposed on how many ellipses we are allowed to fit in one frame, which is fixed at 6 by default. If there are more possible fits available, then we choose the edge combinations with a combined length that is closest to $\hat{C}$.
	
	We fit an ellipse on each edge combination that remains. The properties of every ellipse are subsequently calculated from the general equation of the ellipse using rotation transformation \cite{zhu1999}. This includes the position, semi-major and minor axes, rotation angle and bounding box dimensions. The circumference is computed by Ramanujan’s second approximation \cite{ramanujan1914}. Immediately after obtaining these characteristics, each ellipse passes through a series of filters that will judge the size, shape and quality of the fit. The circumference of the ellipse has to fall within the $C_{min}$ and  $C_{max}$ limits and its aspect ratio should be larger than $AR_{min}$. These bounds have been empirically determined from our data set ($C_{min} = 60$, $C_{max} = 290$ and $AR_{min} = 0.4$). Furthermore, we observe that the limit of $C$ is dependent on $AR$, which is expected because the larger an ellipse is on the surface of a sphere, the greater its polar angle needs to be to have an equally low aspect ratio when projected on the 2D plane as a smaller ellipse. A linear function is created that acts as a circumference threshold, which is given by:
	\begin{equation}
	C'_{max} = k(AR - 1) + C_{max}
	\end{equation}
	Where $k$ is assigned a value of 154. The circumference of the fit has to be less than the circumference limit calculated by this function. Besides looking at the absolute values of the size and shape of the pupil, we also inspect how much these features deviate from the predictions, which is quantified by $\delta_{C}$ and $\delta_{AR}$  calculated through equations \ref{eq:change_C} and \ref{eq:change_AR}.  We establish the criterion that $\delta_{C}$ and $\delta_{AR}$ are not allowed to respectively exceed $\delta_{\theta,C}$ and $\delta_{\theta,AR}$ (see equations \ref{eq:delta_circumference} and \ref{eq:delta_aspect_ratio}), otherwise the fit is rejected. Another filter examines the number of points that the ellipse was fitted on with respect to the fit’s circumference. We cannot accept a large ellipse fit on a tiny set of edge points even when these points are adequately spread out. Such a fit is most likely erroneous and influenced by noisy edges. We therefore enforce the following threshold for the edge length: 
	\begin{equation*}
	L \geq 0.3\hat{C}(1-\delta_{\theta,\hat{C}})
	\end{equation*}
	The last filter works on the basis of the error between the fit and fitted points. For ellipse fits on the pupil-iris edge, we expect a small fit error for all fitted edge points, since we are able to adequately approximate the pupil shape by an ellipse \cite{wyatt1995}. So a large fit error for a particular edge point would indicate that the ellipse is not at all or not entirely fitted on the pupil boundary. The accuracy of the pupil fit can already significantly diminish if the edge combination contains just a few outliers, because of the high sensitivity of direct least squares compared to other, slower, ellipse fitting methods \cite{Halir98}. For this reason, only ellipse fits where all edge points in the set have a small fit error should be accepted. On the other hand, we do not want to reject a fit because of just one outlier. As a compromise, we instead consider the fit errors of the $0.05C$ largest outliers of the set. If their average fit error is above a given threshold, the ellipse fit is rejected.
	
	We also investigate how the absolute fit errors change with respect to the circumference of the ellipse fit by analyzing the fit errors of around 44,000 ellipse fits. Only fits that the pupil detection algorithm has classified as acceptable pupil fits are considered here. We observe that $\epsilon$ is directly proportional to $C$ with larger errors being measured at larger circumferences. Since we do not want to adapt our fit error threshold according to $C$, we remove this linear relationship by using the relative fit error $\epsilon_{rel}$ instead, which is given by:
	\begin{equation*}
	\epsilon_{rel} = \frac{\epsilon_{abs} - \alpha}{C}
	\end{equation*}
	
	Where $\alpha$ (y-intercept) is a constant found by linear regression ($\alpha = -0.56$). The dependence of $\epsilon$ on $C$ disappears when using the relative error. This allows us to set a constant error threshold, which is assigned a value of 0.6.
	
	In many cases, only one ellipse fit remains at this stage which we then consider to represent the pupil. However, it also frequently occurs that there are still a few ellipse fits left to choose from. The final choice is made through a similar strategy employed during edge classification. Each ellipse is assigned a score based on a number of its features, after which the ellipse with the highest score is selected as the pupil representation.  The relevant features are listed in Table \ref{table:fit_features}. Once again, the feature values are made invariant to the image size. The labelled data set was acquired by marking accepted fits as pupil fits and any other ones as non-pupil fits.
	
	\begin{table}[ht]
		\centering
		\resizebox{\textwidth}{!}{%
			\begin{tabular}{ M{3cm} M{2cm} M{3.5cm} M{2.5cm} m{5cm} N }
				
				\textbf{Fit features} 
				& \textbf{Symbol} 
				& \textbf{Feature value, $F_{i}$} 
				& \textbf{Weight factor} 
				& \textbf{Description}
				&\\[20pt]
				
				\hline
				Circumference 
				& $C$
				& $\displaystyle{\frac{\abs{ C - \hat{C} }}{\max(C,\hat{C})}}$ 
				& $c_{A}$ 
				& Circumference of ellipse fit in pixels
				&\\[25pt]
				
				\hline
				Aspect ratio
				& $AR$
				& $\displaystyle{\abs{ AR - \hat{AR} }}$ 
				& $c_{A}$
				& Aspect ratio of ellipse fit  
				&\\[25pt]
				
				\hline
				Edge length 
				& $L$
				& $\displaystyle{\frac{\abs{ L - \hat{L} }}{\max(L,\hat{L})}}$
				& $c_{A}$  
				& Combined length of fitted edges in pixels
				&\\[25pt]
				
				\hline
				Fit error
				& $\epsilon$
				& $\epsilon$
				& - 
				& Relative error between the ellipse fit and the edge points 
				&\\[25pt]
				
				\hline
				Rotation angle
				& $\theta$
				& $\displaystyle{\abs{ \theta - \hat{\theta} }}$ 
				& $c_{A}(1 - \rho(AR))$
				& Rotation angle of ellipse fit in radians 
				&\\[25pt]
				
		\end{tabular}}
		\caption{Ellipse fit features that the fit classification scheme is based on.}
		\label{table:fit_features}
	\end{table}
	
\begin{figure}[ht]
		\centering
		\includegraphics[scale=1.20]{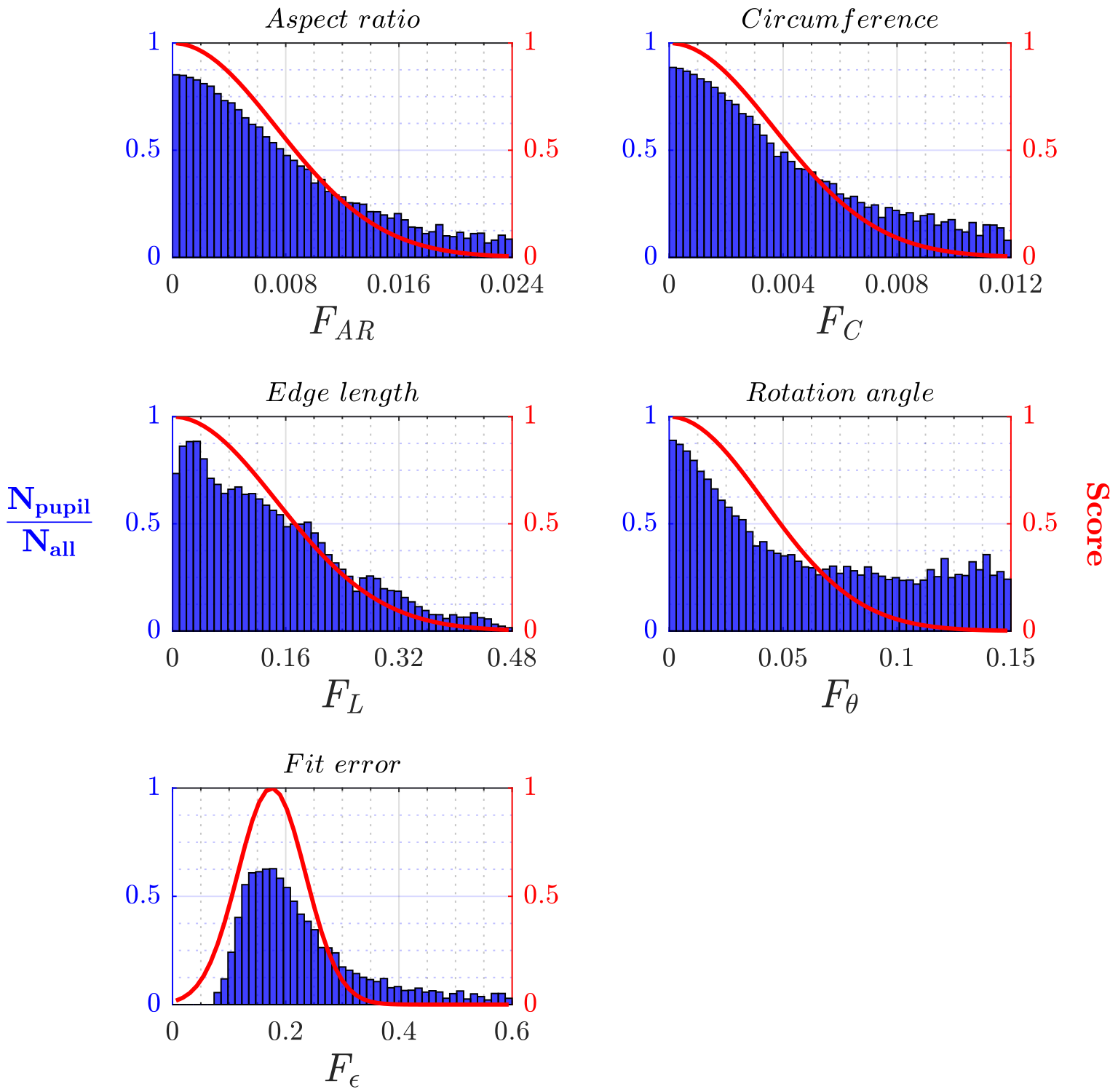}
		\caption{For each of the six feature values given in Table \ref{table:fit_features}, a histogram (blue bars) is plotted with the fraction of ellipse fits that are pupil fits in each bin, similarly to Figure \ref{fig:edge_gaussians}. Gaussian functions have been fitted on the data (red curves), which convert the feature value to a score.}
		\label{fig:fit_gaussians}
	\end{figure}

	Following the same steps as with edge classification, we split our data set into two halves and determine the Gaussian functions and weights from one half, and test the classifier with the other half. The frequency data is plotted in Figure \ref{fig:fit_gaussians}, together with the corresponding Gaussians. The weights are modified by the degree of certainty as indicated in Table \ref{table:fit_features}. An additional factor is added to the rotation angle's weight, because the significance of $\theta$ should decrease the more circular the pupil becomes.
	
\begin{figure}[ht]
		\centering
		\includegraphics[width=0.8\textwidth]{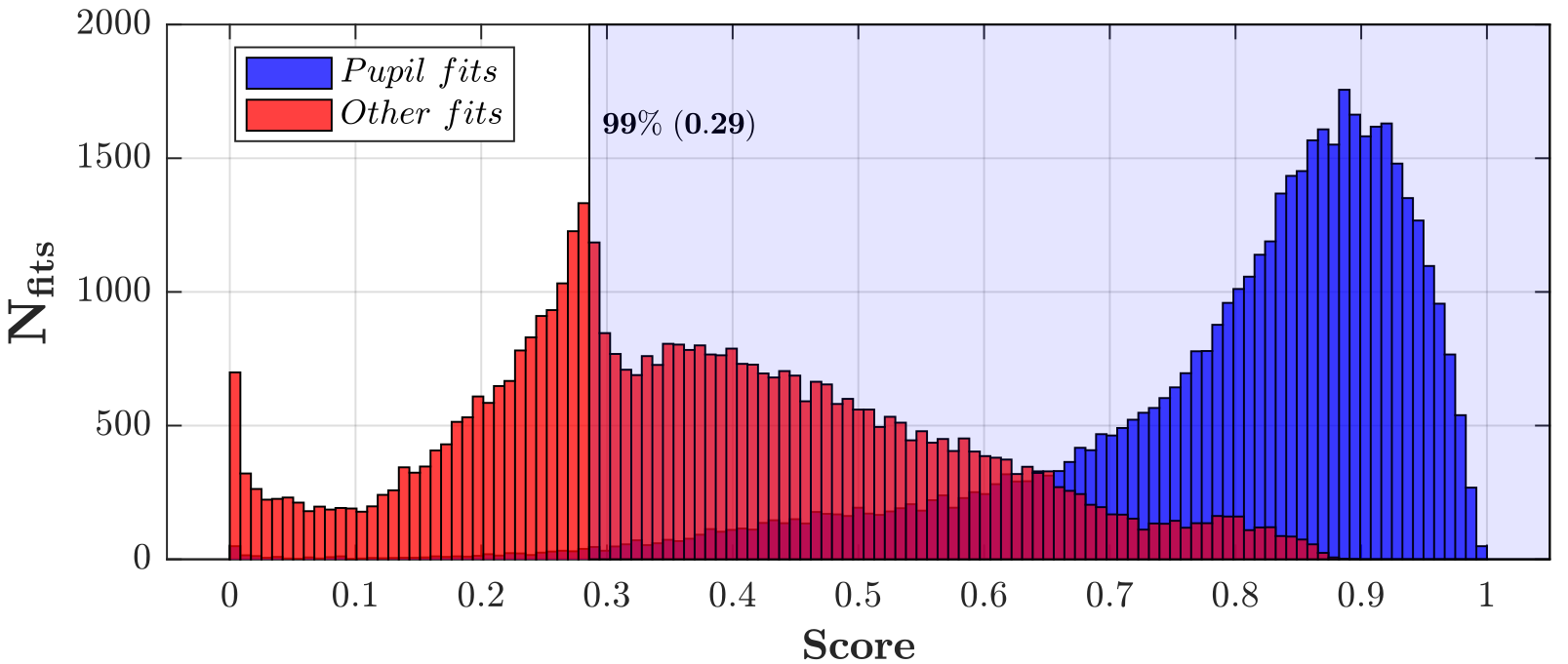}
		\caption{Score histogram of ellipse fits classified as either entirely belonging to the pupil (blue) or not (red). Shown is the greatest separation between the two classes that was obtained with the weights given in Table \ref{table:fit_weights}. The vertical line denotes the score for which 99\% of pupil fits are correctly classified.}
		\label{fig:fit_classifier}
	\end{figure}

	Once more the two-sample Kolmogorov-Smirnov test is used to obtain the greatest degree of separation between the two classes. The optimal weights are given in Table \ref{table:fit_weights}. The distributions of the two classes are plotted in Figure \ref{fig:fit_classifier}. We choose the ellipse fit with the highest score as the pupil representation, but also any fit that has a score within a certain distance $\Delta S_{\theta,fit}$ from the highest score. This parameter $\Delta  S_{\theta,fit}$ is given a value of 0.10. The average characteristics of all accepted fits then corresponds with our final pupil measurement for the current frame.
	
	\begin{table}[ht]
		\centering
		\begin{tabular}{M{1.5cm} M{1.5cm} N}
			
			\textbf{Weight factor} & \textbf{Weight value} \\ 
			
			\hline
			$w_{C}$ 		
			& 0.4 
			&\\[7pt]
			
			\hline
			$w_{AR}$ 		
			& 0.6
			&\\[7pt]
			
			\hline
			$w_{L}$ 		
			& 1.6 
			&\\[7pt]
			
			\hline
			$w_{\epsilon}$ 	
			& 0.9
			&\\[7pt]
			
			\hline
			$w_{\theta}$ 	
			& 1.5
			&\\[7pt]
			
			\hline
			$\rho$			
			& 0.7 
			&\\[7pt]
			
		\end{tabular}
		\caption{Optimal weight factors for maximum separation between the two fit distribution types.}
		\label{table:fit_weights}
	\end{table}
	
	\section{Evaluation}
	
	The performance of the presented pupil detection algorithm, which is given the name EyeStalker, is assessed by determining its effectiveness of locating the pupil centre in images where the position of the pupil has been manually determined. This performance is then compared with two other open-source eye tracking algorithms, PupilLabs \cite{Kassner2014} and ExCuSe \cite{Fuhl2015} (latest versions as of February 2017). A number of different hand-labelled data sets are available \cite{swirski2012} \cite{Kassner2014}, but these are either random collections of eye images or have been recorded with low frame-rate cameras (around 25 Hz). It is a prerequisite for EyeStalker that the algorithm is applied on high frequency image data that is ordered in sequence. For this reason, a new hand-labelled dataset is created that consists of 5000 images of the pupil during 49 saccadic eye movements from 12 different individuals (between 3 to 5 saccades per person), which is a sub-set of the data set referred to in Methods. Especially challenging trials were selected, including images where the pupil is notably obstructed by the eyelid or glint, as well as images of highly eccentric or tiny pupils.
	
	A tool to manually detect the pupil was developed in MATLAB. The program works by fitting an ellipse by hand on the pupil boundary. The centre of the ellipse then corresponds with the centre of the pupil. The ellipse is translated, rotated and reshaped using keyboard controls until a good fit is achieved. The image is resized to fit the entire monitor screen, allowing for more precise measurements. The precision of this method is determined by manually detecting the pupil in 12 images 12 times and calculating the variation in the measurement. Each of these images shows the pupil of a different person and was hand-picked for their higher detection difficulty. The 144 detections are done in a random order and the shape, angle and position of the ellipse is reset each time a new image is displayed on screen. From this analysis, a standard deviation of \SI{0.6}{\degree} is found for manual detection. This labelling tool together with the raw eye images and hand-labelled pupil coordinates are publicly available for download.\footnote{\url{https://drive.google.com/open?id=0Bw57olSwQ4EbUWV5ajNKeG93NEk}}
	
	The hand-labelled data set only features frames that are part of a saccadic eye movement plus a short fixation period before and after. On average, this comes down to around 100 frames per trial. However, when applying the three algorithms on the data, we include a longer preceding fixation period, adding between 20 to 200 additional frames to the start of the trial, which are not included in our analysis. These extra prior detections are required for EyeStalker to already acquire an adequately accurate measure of $\bar{f}_{n}$ before the actual measurement begins. In normal circumstances, more accurate information will always be available, because a test subject will have been wearing the eye tracking device for a relatively long time before the start of a recording.
	
\begin{figure}[ht]
		\centering
		\includegraphics[scale=1.00]{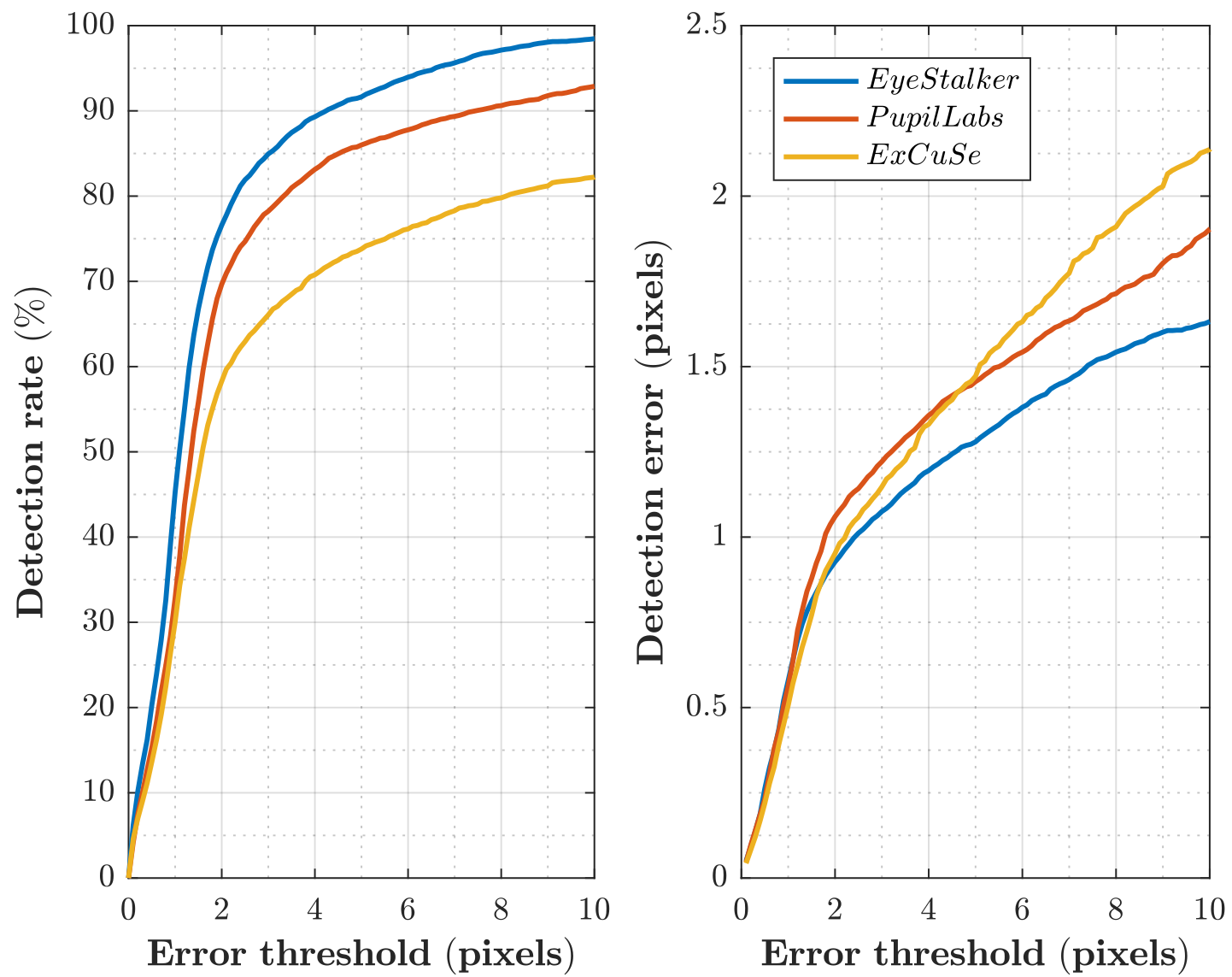}
		\caption{Performance of our pupil detection algorithm (EyeStalker) on 5000 test images compared to two other open-source algorithms, PupilLabs and ExCuSe. The detection rate and error are evaluated as a function of the allowable error threshold. EyeStalker not only achieves a significantly higher detection rate, but is also able to achieve this with an overall greater accuracy.}
		\label{fig:p1_results_performance}
	\end{figure}

\begin{figure}[ht]
		\centering
		\includegraphics[scale=1.00]{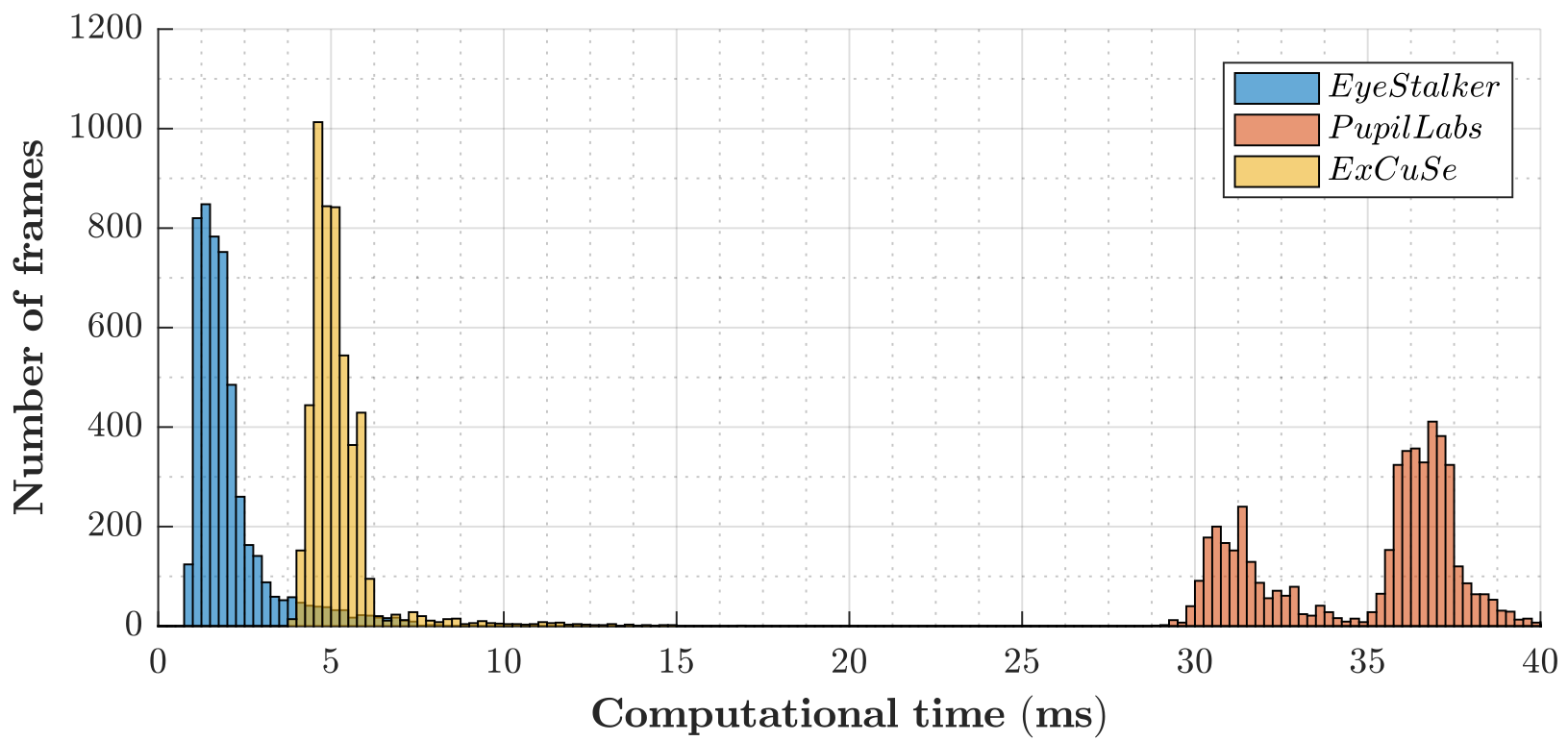}
		\caption{Histograms of the processing time for each of the 5000 test images are plotted for the three algorithms. Our pupil detection algorithm (EyeStalker) achieves an average computation time of 2.1 ms per frame.}
		\label{fig:p1_results_speed}
	\end{figure}

	As a constraint, all trials are processed in a single run, which means that parameters are kept to their default values and not altered between individuals. In the EyeStalker and PupilLabs algorithms, the parameters for Canny edge detection are set equal to one another to achieve a fairer comparison (ExCuSe uses an automatized method to set these parameters). The performance of each algorithm is evaluated by calculating the detection rate and error for different error thresholds. The results are shown in Figure \ref{fig:p1_results_performance}. It is clear that EyeStalker not only achieves a greater detection rate for all error thresholds compared to the other two approaches, but also a smaller detection error. The much lower detection rate of ExCuSe can possibly be attributed to the fact that the algorithm is specifically tailored for eye tracking in real-life scenarios and thus functions relatively poorly in controlled environments.
	The performance of PupilLabs comes closer to that of EyeStalker, but requires considerably more processing time, which can be inferred from Figure \ref{fig:p1_results_speed}. The figure shows detection durations for every frame, which were processed using a C++ implementation of each algorithm combined in a single application on a triple-core 3.3 GHz CPU running Linux. EyeStalker is significantly faster than both of the other two algorithms. PupilLabs is surprisingly slow, but the timing agrees with their reported processing pipeline latency of 45 ms \cite{Kassner2014}, although it is also possible that our implementation or hardware was suboptimal, since their commercially available hardware reportedly has a latency of 5.7 ms.\footnote{\url{https://pupil-labs.com/}} The average computational time of EyeStalker is 2.1 ms, but this duration can potentially double during periods of low certainty. The vast majority of processing time is spent on approximate detection (section \ref{sec:approximate_detection}) and Canny edge detection (section \ref{sec:canny_edge_detection}).
	
	\section{Discussion}
	
	A novel eye tracking algorithm has been presented that is designed for use with high-speed cameras, relying on estimations of pupil characteristics to carry out sophisticated feature-based pupil detection techniques that are both fast and robust. Based on ground truth hand-labelled data, the method was found to surpass other recently published open-source algorithms in terms of detection rate, accuracy and speed. To further verify the efficiency of the algorithm, it should be tested with multiple camera set-ups using different combinations of operational speed and image quality. It is expected that performance will increase with higher frame rate cameras since the pupil feature predictions will become more accurate.
	
	Additional enhancements can be made to the algorithm itself. The implemented recursive estimation method is sufficiently accurate, but using a more established design (e.g. Kalman filter) will most likely lead to better predictions, though requiring more effort to set-up. This could significantly improve pupil detection, since the predictions are used in many different parts of the algorithm. One processing step that does not use any predictions, however, is Canny edge detection, but the estimated pupil location can be utilized to tune the method to specifically detect the pupil-iris edge. In its current state, Canny edge detection does not make any distinctions between edges based on their orientation or gradient direction. It can be made more discerning by calculating the radial gradient outwards from the predicted pupil position, which puts more weight on the pupil contour and less on edges that are not oriented in the tangential direction (e.g. eye lashes). Furthermore, by using the signed gradient, we only consider edge points that are darker closer to the origin point and lighter on the other side, which holds true for the pupil perimeter. An additional improvement can be made in the approximate detection of the pupil position. Even though Haar-like feature detection is made significantly more computationally efficient with the calculation of the integral image compared to more naive methods, it is still quite demanding relative to other processing steps. Since we only require an approximate pupil location, an alternative approach is to train a convolution neural network on the eye image data, which would subsequently be able to rapidly supply a rough position estimate in each frame.
	
	\bibliography{bibfile}	

\begin{thebibliography}{10}

\bibitem{abbott2012}
W~W Abbott and A~A Faisal.
\newblock Ultra-low-cost 3d gaze estimation: an intuitive high information
  throughput compliment to direct brain–machine interfaces.
\newblock {\em Journal of Neural Engineering}, 9(4):046016, 2012.

\bibitem{abd2002}
Wael Abd-Almageed, M~Sami Fadali, and George Bebis.
\newblock A non-intrusive kalman filter-based tracker for pursuit eye movement.
\newblock In {\em American Control Conference, 2002. Proceedings of the 2002},
  volume~2, pages 1443--1447. IEEE, 2002.

\bibitem{adiba2016}
A.~I. Adiba, N.~Tanaka, and J.~Miyake.
\newblock An adjustable gaze tracking system and its application for automatic
  discrimination of interest objects.
\newblock {\em IEEE/ASME Transactions on Mechatronics}, 21(2):973--979, April
  2016.

\bibitem{allison1996}
Robert~S Allison, Moshe Eizenman, and Bob~SK Cheung.
\newblock Combined head and eye tracking system for dynamic testing of the
  vestibular system.
\newblock {\em IEEE Transactions on Biomedical Engineering}, 43(11):1073--1082,
  1996.

\bibitem{opencv_library}
G.~Bradski.
\newblock The opencv library.
\newblock {\em Dr. Dobb's Journal of Software Tools}, 2000.

\bibitem{canny1986}
J.~Canny.
\newblock A computational approach to edge detection.
\newblock {\em IEEE Transactions on Pattern Analysis and Machine Intelligence},
  PAMI-8(6):679--698, Nov 1986.

\bibitem{chi2014}
Jian-nan Chi, Li-hua Xie, Peng-yun Zhang, Yi-fang Lu, and Guo-sheng Zhang.
\newblock Hybrid particle and kalman filtering for pupil tracking in active ir
  illumination gaze tracking system.
\newblock {\em Mathematical Problems in Engineering}, 2014, 2014.

\bibitem{Cormen2001}
Thomas~H. Cormen, Clifford Stein, Ronald~L. Rivest, and Charles~E. Leiserson.
\newblock {\em Introduction to Algorithms}.
\newblock McGraw-Hill Higher Education, 2nd edition, 2001.
\newblock p. 978.

\bibitem{dolezal2015}
J.~Doležal and V.~Fabian.
\newblock 41. application of eye tracking in neuroscience.
\newblock {\em Clinical Neurophysiology}, 126(3):e44 --, 2015.

\bibitem{duchowski2002}
Andrew~T Duchowski.
\newblock A breadth-first survey of eye-tracking applications.
\newblock {\em Behavior Research Methods, Instruments, \& Computers},
  34(4):455--470, 2002.

\bibitem{fitzgibbon1999}
Andrew Fitzgibbon, Maurizio Pilu, and Robert~B. Fisher.
\newblock Direct least square fitting of ellipses.
\newblock {\em IEEE Trans. Pattern Anal. Mach. Intell.}, 21(5):476--480, May
  1999.

\bibitem{Fuhl2015}
Wolfgang Fuhl, Thomas K{\"u}bler, Katrin Sippel, Wolfgang Rosenstiel, and
  Enkelejda Kasneci.
\newblock {\em ExCuSe: Robust Pupil Detection in Real-World Scenarios}, pages
  39--51.
\newblock Springer International Publishing, Cham, 2015.

\bibitem{Halir98}
Radim Halir and Jan Flusser.
\newblock Numerically stable direct least squares fitting of ellipses, 1998.

\bibitem{heide1999}
W.~Heide, E.~Koenig, P.~Trillenberg, D.~Kömpf, and {D. S.} Zee.
\newblock Electrooculography: technical standards and applications. the
  international federation of clinical neurophysiology.
\newblock {\em Electroencephalography and clinical neurophysiology.
  Supplement}, 52:223--240, 1999.

\bibitem{Kassner2014}
Moritz Kassner, William Patera, and Andreas Bulling.
\newblock Pupil: An open source platform for pervasive eye tracking and mobile
  gaze-based interaction.
\newblock In {\em Adjunct Proceedings of the 2014 ACM International Joint
  Conference on Pervasive and Ubiquitous Computing}, UbiComp '14 Adjunct, pages
  1151--1160, New York, NY, USA, 2014. ACM.

\bibitem{kim2014}
Elizabeth~S Kim, Adam Naples, Giuliana~Vaccarino Gearty, Quan Wang, Seth
  Wallace, Carla Wall, Michael Perlmutter, Jennifer Kowitt, Linda Friedlaender,
  Brian Reichow, et~al.
\newblock Development of an untethered, mobile, low-cost head-mounted eye
  tracker.
\newblock In {\em Proceedings of the Symposium on Eye Tracking Research and
  Applications}, pages 247--250. ACM, 2014.

\bibitem{li2006}
Dongheng Li, Jason Babcock, and Derrick~J Parkhurst.
\newblock openeyes: a low-cost head-mounted eye-tracking solution.
\newblock In {\em Proceedings of the 2006 symposium on Eye tracking research \&
  applications}, pages 95--100. ACM, 2006.

\bibitem{li2005}
Dongheng Li, D.~Winfield, and D.~J. Parkhurst.
\newblock Starburst: A hybrid algorithm for video-based eye tracking combining
  feature-based and model-based approaches.
\newblock In {\em 2005 IEEE Computer Society Conference on Computer Vision and
  Pattern Recognition (CVPR'05) - Workshops}, pages 79--79, June 2005.

\bibitem{lin2010}
Lin Lin, Lin Pan, LiFang Wei, and Lun Yu.
\newblock A robust and accurate detection of pupil images.
\newblock In {\em Biomedical Engineering and Informatics (BMEI), 2010 3rd
  International Conference on}, volume~1, pages 70--74. IEEE, 2010.

\bibitem{long2007}
Xindian Long, Ozan~K Tonguz, and Alex Kiderman.
\newblock A high speed eye tracking system with robust pupil center estimation
  algorithm.
\newblock In {\em Engineering in Medicine and Biology Society, 2007. EMBS 2007.
  29th Annual International Conference of the IEEE}, pages 3331--3334. IEEE,
  2007.

\bibitem{maini2009}
Raman Maini and Himanshu Aggarwal.
\newblock Study and comparison of various image edge detection techniques.
\newblock {\em International journal of image processing (IJIP)}, 3(1):1--11,
  2009.

\bibitem{mantiuk2012}
Rados{\l}aw Mantiuk, Micha{\l} Kowalik, Adam Nowosielski, and Bartosz Bazyluk.
\newblock Do-it-yourself eye tracker: Low-cost pupil-based eye tracker for
  computer graphics applications.
\newblock In {\em International Conference on Multimedia Modeling}, pages
  115--125. Springer, 2012.

\bibitem{morimoto2000}
Carlos~Hitoshi Morimoto, Dave Koons, Arnon Amir, and Myron Flickner.
\newblock Pupil detection and tracking using multiple light sources.
\newblock {\em Image and vision computing}, 18(4):331--335, 2000.

\bibitem{Newman2010}
Mark Newman.
\newblock {\em Networks: An Introduction}.
\newblock Oxford University Press, Inc., New York, NY, USA, 2010.

\bibitem{puatruaucean2012}
Viorica P{\u{a}}tr{\u{a}}ucean.
\newblock {\em Detection and identification of elliptical structure
  arrangements in images: Theory and algorithms}.
\newblock PhD thesis, INPT, 2012.

\bibitem{patrikalakis2009}
Nicholas~M Patrikalakis and Takashi Maekawa.
\newblock {\em Shape interrogation for computer aided design and
  manufacturing}.
\newblock Springer Science \& Business Media, 2009.
\newblock p. 40.

\bibitem{putra2013}
I~Ketut Gede~Darma Putra, Agung Cahyawan, and Yandi Perdana.
\newblock Low-cost based eye tracking and eye gaze estimation.
\newblock {\em TELKOMNIKA (Telecommunication Computing Electronics and
  Control)}, 9(2):377--386, 2013.

\bibitem{raguram2008}
Rahul Raguram, Jan-Michael Frahm, and Marc Pollefeys.
\newblock A comparative analysis of ransac techniques leading to adaptive
  real-time random sample consensus.
\newblock {\em Computer Vision--ECCV 2008}, pages 500--513, 2008.

\bibitem{ramanujan1914}
Srinivasa Ramanujan.
\newblock Modular equations and approximations to $ \pi $.
\newblock {\em Quarterly Journal of Mathematics}, 45:180, 350--372, 1914.

\bibitem{schneider2011}
Nicolas Schneider, Peter Bex, Erhardt Barth, and Michael Dorr.
\newblock An open-source low-cost eye-tracking system for portable real-time
  and offline tracking.
\newblock In {\em Proceedings of the 1st Conference on Novel gaze-controlled
  applications}, page~8. ACM, 2011.

\bibitem{swirski2012}
Lech \'Swirski, Andreas Bulling, and Neil~A. Dodgson.
\newblock Robust real-time pupil tracking in highly off-axis images.
\newblock In {\em Proceedings of ETRA}, March 2012.

\bibitem{geest2002}
JN~Van~der Geest and MA~Frens.
\newblock Recording eye movements with video-oculography and scleral search
  coils: a direct comparison of two methods.
\newblock {\em Journal of neuroscience methods}, 114(2):185--195, 2002.

\bibitem{viola2001}
Paul Viola and Michael Jones.
\newblock Robust real-time object detection.
\newblock In {\em International Journal of Computer Vision}, 2001.

\bibitem{volck2015}
Alexander~C Volck, Roman~D Laske, Ralph Litschel, Rudolf Probst, and Abel-Jan
  Tasman.
\newblock Sound localization measured by eye-tracking.
\newblock {\em International journal of audiology}, 54(12):976--983, 2015.

\bibitem{welch95}
Greg Welch and Gary Bishop.
\newblock An introduction to the kalman filter, 1995.

\bibitem{wren1997}
Christopher~Richard Wren, Ali Azarbayejani, Trevor Darrell, and Alex~Paul
  Pentland.
\newblock Pfinder: Real-time tracking of the human body.
\newblock {\em IEEE Transactions on pattern analysis and machine intelligence},
  19(7):780--785, 1997.

\bibitem{wyatt1995}
Harry~J Wyatt.
\newblock The form of the human pupil.
\newblock {\em Vision Research}, 35(14):2021--2036, 1995.

\bibitem{zhang2006}
Jiashu Zhang and Zutao Zhang.
\newblock Application of a strong tracking finite-difference extended kalman
  filter to eye tracking.
\newblock In {\em International Conference on Intelligent Computing}, pages
  1170--1179. Springer, 2006.

\bibitem{zhu1999}
Danjie Zhu, Steven~T. Moore, and Theodore Raphan.
\newblock Robust pupil center detection using a curvature algorithm.
\newblock {\em Computer Methods and Programs in Biomedicine}, 59(3):145 -- 157,
  1999.

\bibitem{zimmermann2016}
Jan Zimmermann, Yuriria Vazquez, Paul~W. Glimcher, Bijan Pesaran, and Kenway
  Louie.
\newblock Oculomatic: High speed, reliable, and accurate open-source eye
  tracking for humans and non-human primates.
\newblock {\em Journal of Neuroscience Methods}, 270:138 -- 146, 2016.

\end{thebibliography}
	
\end{document}